\documentclass{article}

\usepackage{microtype}
\usepackage{graphicx}
\usepackage{subfigure}
\usepackage{booktabs} %

\usepackage[accepted]{mlsys2025}

\usepackage{amsmath,amssymb,amsfonts}
\usepackage{graphicx}
\usepackage{textcomp}
\usepackage{xcolor}
\usepackage{multirow}
\usepackage{booktabs}
\usepackage{pifont}
\usepackage{array}
\usepackage{xspace}
\usepackage{soul}
\usepackage{subcaption}
\usepackage{fancyhdr}
\usepackage{colortbl}
\usepackage{makecell}
\usepackage[bookmarks=true,breaklinks=true,letterpaper=true,colorlinks,citecolor=blue,linkcolor=blue,urlcolor=blue]{hyperref}
\usepackage{enumitem}

\usepackage{algorithm}
\usepackage{algpseudocode}
\usepackage{amsmath}
\usepackage{amsfonts}

\usepackage{hyperref}
\usepackage{caption}
\usepackage{subcaption}

\usepackage{microtype}
\usepackage{subfigure}

\usepackage[utf8]{inputenc} %
\usepackage{inconsolata} %
\usepackage[most]{tcolorbox} %
\usepackage{soul} %
\usepackage{siunitx} %
\usepackage{pifont} %
\usepackage{xurl} %
\usepackage[textsize=scriptsize, linecolor=magenta, bordercolor=magenta,
  backgroundcolor=white, textwidth=45pt]{todonotes}
\usepackage{marginnote}

\usepackage[skip=3pt, font=small]{caption}
\newcommand{\sect}[1]{Section~\ref{#1}}

\def\naive{na\"{\i}ve\xspace}

\makeatletter
\DeclareRobustCommand\onedot{\futurelet\@let@token\@onedot}
\def\@onedot{\ifx\@let@token.\else.\null\fi\xspace}

\makeatother

\definecolor{mydarkblue}{rgb}{0,0.08,1}
\definecolor{mydarkgreen}{rgb}{0.02,0.6,0.02}
\definecolor{mydarkred}{rgb}{0.8,0.02,0.02}
\definecolor{mydarkorange}{rgb}{0.40,0.2,0.02}
\definecolor{mydarkgold}{rgb}{0.6, 0.4, 0.05}
\definecolor{myred}{rgb}{1.0,0.0,0.0}
\definecolor{mygold}{rgb}{0.75,0.6,0.12}
\definecolor{myblue}{rgb}{0,0.2,0.8}
\definecolor{mylightblue}{rgb}{0.827,0.937,0.945}
\definecolor{mydarkgray}{rgb}{0.66,0.66,0.66}
\definecolor{mygray}{rgb}{0.85,0.85,0.85}

\def\system{LServe\xspace}

\renewcommand{\algorithmicrequire}{\textbf{Input:}}
\renewcommand{\algorithmicensure}{\textbf{Output:}}
\newcommand{\myparagraph}[1]{\noindent\textbf{#1}}

\usepackage{listings}
\usepackage{xcolor}
\definecolor{LightGray}{gray}{0.98}
\definecolor{codegreen}{rgb}{0,0.6,0}
\definecolor{codegray}{rgb}{0.5,0.5,0.5}
\definecolor{codepurple}{rgb}{0.58,0,0.82}
\definecolor{backcolour}{rgb}{0.95,0.95,0.92}

\lstdefinestyle{mystyle}{
    frame=single,                     
    framesep=1mm,                     
    framerule=0.4pt,                  
    rulecolor=\color{black!80},       
    xleftmargin=5pt,                  
    xrightmargin=5pt,                     backgroundcolor=\color{backcolour},
    tabsize=2,                        
    breaklines=true,                  
    breakatwhitespace=false,          
    keepspaces=true,                  
    basicstyle=\ttfamily\footnotesize,
    numbers=left,                     
    numbersep=5pt,                    
    numberstyle=\footnotesize\color{codegray}, 
    commentstyle=\color{codegreen},   
    keywordstyle=\color{magenta},         stringstyle=\color{codepurple},   
    identifierstyle=\color{black},    
    showspaces=false,                 
    showstringspaces=false,           
    showtabs=false,                   
    captionpos=b                      
}

\usepackage{fancyhdr} 
\mlsystitlerunning{LServe: Efficient Long-sequence LLM Serving with Unified Sparse Attention}

\usepackage{graphicx}
\usepackage{hyperref}
\usepackage{eso-pic}
\usepackage{xparse}

\newlength{\badgewidth}
\newlength{\badgegap}
\newcommand{\badgeList}{}

\NewDocumentCommand{\addTopRightBadge}{O{} m}{%
\gappto{\badgeList}{\href{#1}{\includegraphics[width=\badgewidth]{#2}}\hspace{\badgegap}}%
}

\newcommand{\placeTopRightBadges}{%
\AddToShipoutPictureBG*{%
\put(\LenToUnit{\paperwidth - 1.5cm - \badgewidth},\LenToUnit{\paperheight - 2cm}){%
\makebox[0pt][r]{\badgeList}%
}%
}%
}

\addTopRightBadge{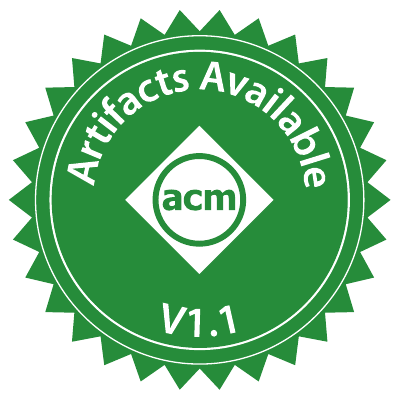}
\addTopRightBadge{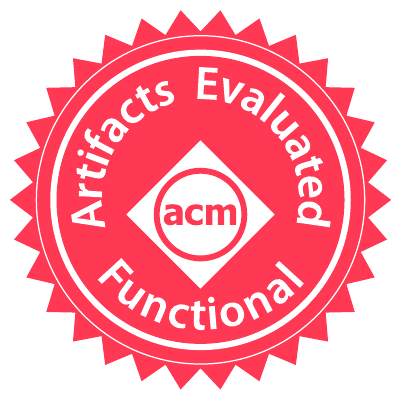}

\placeTopRightBadges

\begin{document}

\twocolumn[

\mlsystitle{LServe: Efficient Long-sequence LLM Serving with \\ Unified Sparse Attention}

\mlsyssetsymbol{equal}{*}

\begin{mlsysauthorlist}
\mlsysauthor{Shang Yang}{equal,mit}
\mlsysauthor{Junxian Guo}{equal,mit,sjtu}
\mlsysauthor{Haotian Tang}{mit}
\mlsysauthor{Qinghao Hu}{mit}

\mlsysauthor{Guangxuan Xiao}{mit}
\mlsysauthor{Jiaming Tang}{mit}
\mlsysauthor{Yujun Lin}{mit}
\mlsysauthor{Zhijian Liu}{nv}
\mlsysauthor{Yao Lu}{nv}
\mlsysauthor{Song Han}{mit,nv}
\end{mlsysauthorlist}

 \mlsyscorrespondingauthor{Song Han}{songhan@mit.edu}
\mlsysaffiliation{mit}{MIT}
\mlsysaffiliation{nv}{NVIDIA}
\mlsysaffiliation{sjtu}{SJTU}

\begin{center}
\url{https://hanlab.mit.edu/projects/lserve} 
\end{center}

\mlsyskeywords{Machine Learning, MLSys}

\vskip 0.15in

\begin{abstract}
Large language models (LLMs) have shown remarkable potential in processing long sequences and complex reasoning tasks, yet efficiently serving these models remains challenging due to the quadratic computational complexity of attention in the prefilling stage and the large memory footprint of the KV cache in the decoding stage. To address these issues, we introduce \system, an efficient system that accelerates long-sequence LLM serving via hybrid sparse attention. This method unifies different hardware-friendly, structured sparsity patterns for both prefilling and decoding attention into a single framework, where computations on less important tokens are skipped block-wise. \system demonstrates the compatibility of static and dynamic sparsity in long-context LLM attention. This design enables multiplicative speedups by combining these optimizations. Specifically, we convert half of the attention heads to nearly free streaming heads in both the prefilling and decoding stages. Additionally, we find that only a constant number of KV pages is required to preserve long-context and reasoning capabilities, irrespective of context length. We then design a hierarchical KV page selection policy that dynamically prunes KV pages based on query-centric similarity. On average, \system accelerates LLM prefilling by up to \textbf{2.9$\times$} and decoding by \textbf{1.3-2.1$\times$} over vLLM, maintaining long-context accuracy. Code is released at \url{https://github.com/mit-han-lab/omniserve}.

\end{abstract}

]

\printAffiliationsAndNotice{\mlsysEqualContribution}  %

\section{Introduction}
\label{sect:intro}

Large Language Models (LLMs) have dramatically transformed the field of artificial intelligence. With expanding context window lengths and test-time computing~\cite{openai2024o1, snell2024scaling}, LLMs now demonstrate remarkable performance across diverse long-sequence applications~\cite{google2024gemini}, including multi-turn conversations, long document analysis~\cite{zhang2024benchmarking, goyal2020evaluating, huang2021efficient}, multi-modal understanding~\cite{xue2024longvila, liu2024visual, lin2024vila}, code completion~\cite{li2023starcoder, lozhkov2024starcoder}, and complex reasoning tasks ~\cite{deepseekai2025deepseekr1incentivizingreasoningcapability}. Many of these applications require processing hundreds of thousands of context tokens in real-world settings, presenting unique challenges. In particular, the demand for fast prefilling, or minimizing the time to the first token, and the burden on the decoding phase due to the large KV (key-value) caches necessary for such contexts, represent significant hurdles.

\begin{figure}[t]
    \centering
    \includegraphics[width=\linewidth]{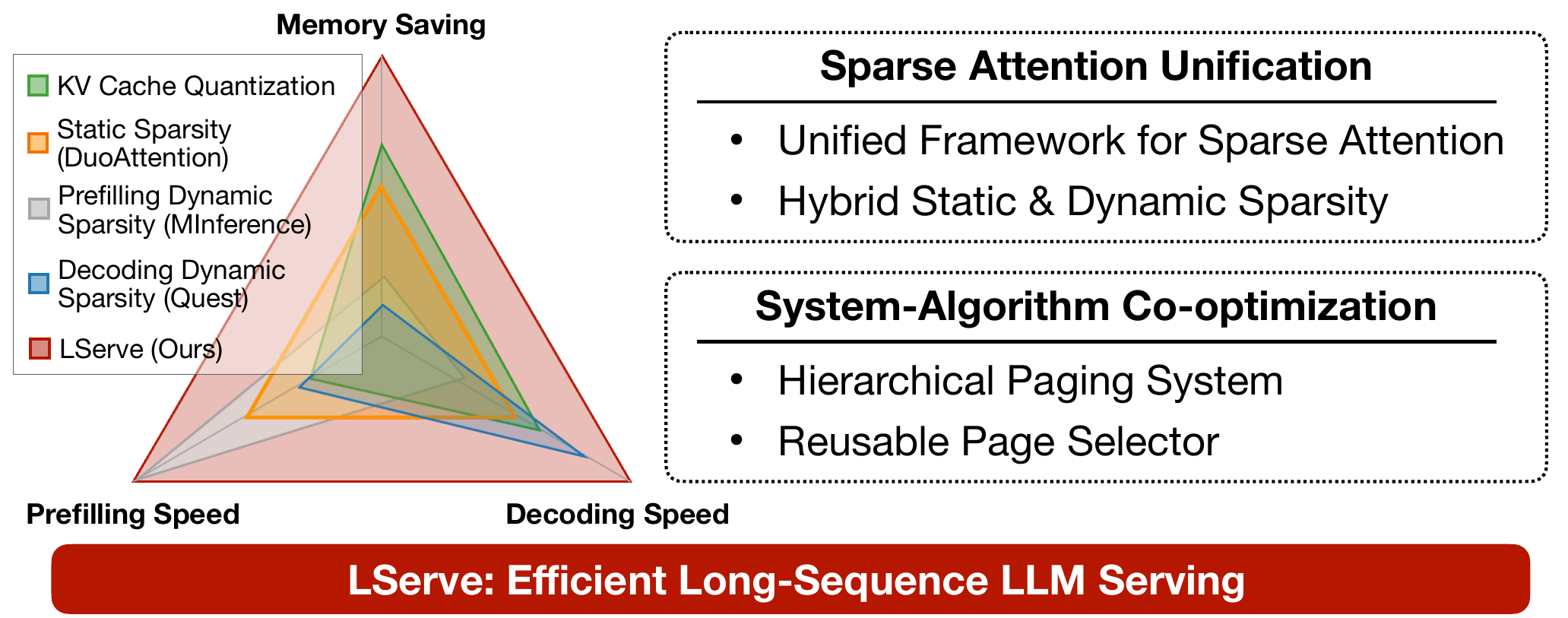}
    \caption{\system is an efficient system for serving long-sequence LLMs that leverages hybrid sparse attention. With the unification of different sparse patterns as well as KV cache quantization, \system achieves significant speedups in both prefilling stage and decoding stage while also reducing the memory consumption.} %
    \label{fig:intro:teaser}
\end{figure}

Long-sequence LLMs are about more than just an extended context. The recently announced OpenAI o1~\cite{openai2024o1} demonstrates exceptional capabilities in complex reasoning tasks, such as deciphering, mathematics, and coding. These advancements are achieved through inference-time scaling and the generation of extensive chains of thought~\cite{wei2022chain}. According to \citet{qin2024o1}, o1's internal reasoning process can extend to \textbf{20k} tokens for mathematical problems, making it the first known \textit{long-generation} LLM. Contrary to the conventional belief that the prefilling stage dominates runtime in long-sequence LLMs, when Llama-3-8B~\cite{dubey2024llama} is run using TensorRT-LLM~\cite{trtllm} with 256k input tokens and 20k output tokens (comparable to o1’s reasoning trace length), the prefilling time is 116 seconds, while decoding takes 540 seconds — almost \textbf{5$\times$} longer.

To enhance the efficiency of long-sequence LLMs, it is essential to optimize both the prefilling and decoding stages rather than focusing on just one. Beyond model architectural modifications in the pre-training stage~\cite{ainslie2023gqa, brandon2024reducing}, existing acceleration solutions for long-sequence LLMs primarily address efficiency from two angles. The first approach centers on KV cache quantization, where methods such as QServe~\cite{lin2024qserve}, KIVI~\cite{liu2024kivi}, and KVQuant~\cite{hooper2024kvquant} employ low-bit quantization to reduce memory usage and I/O traffic, potentially increasing generation throughput. However, these quantization techniques do not lower the number of computations performed in the attention loop, resulting in suboptimal generation speeds as sequence lengths grow. The second approach utilizes approximate sparse attention to improve long-sequence LLM performance. For example, StreamingLLM~\cite{xiao2023efficient}, H2O~\cite{zhang2024h2o}, and TOVA~\cite{oren2024transformers} apply static masking mechanisms to reduce attention complexity, though at the expense of accuracy in long-context tasks and irregular KV cache memory layouts. DuoAttention~\cite{xiao2024duoattention} advances this strategy by pruning attention computations at a coarser granularity using an optimization-based approach. Other methods, such as MInference~\cite{jiang2024minference} and Quest~\cite{tang2024quest}, implement dynamic sparse attention to accelerate either the prefilling \textbf{\textit{or}} decoding stage. However, these approaches do not reduce KV cache memory consumption and lack a unified framework to address efficiency challenges in \textbf{\textit{both}} stages simultaneously.

To this end, we introduce \textbf{\system}, an efficient system for serving long-sequence LLMs that leverages hybrid sparse attention. Recognizing that not all tokens hold equal importance, \system integrates multiple hardware-friendly, structured sparsity patterns into a \textbf{unified block sparse attention} framework (see Figure~\ref{fig:method:blocksparse}). Block-level sparsity accelerates attention computation by processing the KV history in discrete blocks. By skipping blocks, we directly reduce the number of sequential iterations, resulting in measured speedups during both the prefilling and decoding stages.

Building on the unified block sparse attention framework, \system further illustrates acceleration opportunities from \textit{static} and \textit{dynamic} sparsity. 

For \textit{static} sparsity, inspired by DuoAttention~\cite{xiao2024duoattention}, we modify the attention masks in the original model by converting half of the attention heads into $\Lambda$-shaped masks, transforming these attention heads into \textit{streaming heads}. Additionally, we fuse the computation of streaming and standard attention heads into unified GPU kernels for both the prefilling and decoding stages, translating theoretical computation and memory savings that translate to up to 1.7$\times$ measured speedup. 

For \textit{dynamic} sparsity, we observe that query-centric sparsity~\cite{tang2024quest} allows for nearly lossless KV compression: the required number of KV tokens to maintain long-context capabilities remains constant (e.g., 4096), regardless of context length. To optimize efficiency, we design a hierarchical page selector to identify important KV pages for each query token, reusing the selection results across tokens to reduce page selection overhead by 4$\times$. 

Our key observation is that \textbf{static and dynamic sparsity patterns are orthogonal} in long-sequence LLMs. By unifying static and dynamic sparsity with KV cache quantization into a single GPU kernel, \system achieves compounded efficiency benefits from each individual optimization for decoding stage attention.

We benchmark \system across three long-sequence LLMs—Llama-3-8B, Minitron-4B, and Llama-2-7B—at context lengths up to 512k tokens. Compared to state-of-the-art frameworks like vLLM~\cite{vllm}, QServe~\cite{lin2024qserve}, MInference~\cite{jiang2024minference}, and DuoAttention~\cite{xiao2024duoattention}, \system accelerates prefilling stage by up to \textbf{2.9$\times$} and achieves an average of \textbf{1.3$\times$-2.1$\times$} speedup in the decoding stage, without sacrificing the long-context capabilities of the original dense, floating-point models. Furthermore, \system also demonstrates compatibility with the latest reasoning-centric models. Table~\ref{tab:results:reasoning} indicates that \system achieves comparable accuracy to dense baselines on DeepSeek-R1-Distill-Llama-8B~\cite{deepseekai2025deepseekr1incentivizingreasoningcapability}.

\section{Background and Motivation}
\label{sect:background}

\subsection{Background}

\paragraph{LLM Inference.} LLMs are transformer-based architectures with stacked identical layers, each containing attention blocks, feed-forward networks (FFN), and normalization components. %
LLM inference involves two stages: an initial \emph{prefilling} stage that handles multiple tokens concurrently, followed by auto-regressive \emph{decoding} stage where only one token will be processed for each request in a decoding step.

\myparagraph{Attention.} The attention mechanism exchanges information across tokens. It first transforms input $\mathbf{x}$ through linear projections to generate query vectors $\mathbf{q}\in\mathbb{R}^{N\times HD}$, and key-value pairs $\mathbf{k},\mathbf{v}\in\mathbb{R}^{N\times \hat{H}D}$, where $\hat{H}$ represents the key/value head count. Traditional multi-head attention (MHA) maintains $H = \hat{H}$, and contemporary architectures 
~\cite{touvron2023llama2,jiang2023mistral,jiang2024mixtral} employ grouped-query attention (GQA)~\cite{ainslie2023gqa}
where $H = n\hat{H} (n\in \mathbb{Z})$ to shrink the size of KV cache. The current $\mathbf{k}$ and $\mathbf{v}$ is then concatenated with KV cache from $S$ preceding tokens, yielding $\mathbf{K}, \mathbf{V}\in\mathbb{R}^{(S+N)\times \hat{H}D}$. The attention computation can then be formulated as follows:

\vspace{-14pt}
\begin{equation}
\small
\mathbf{S}_{h} = \frac{\mathbf{q}_{h}\mathbf{K}_{\hat{h}}^T}{\sqrt{D}}, 
\hspace{5pt}
\mathbf{o}_{h} = \text{softmax}\left(\mathbf{S}_{h}\right)\mathbf{V}_{\hat{h}}, 
\hspace{5pt}
\hat{h}=\left\lfloor\frac{h}{n}\right\rfloor
\label{eqn:background:attn}
\end{equation}

Therefore, the complexity of attention can be expressed as $O\left(N(S+N)HD\right)$, which increases quadratically in the prefilling stage and linearly in the decoding stage with respect to sequence length. When $S$ is long, both decoding stage and prefilling stage are bounded by attention.

\myparagraph{Paged Attention.} In LLM serving, the generation length of each sequence is highly variable and unpredictable. Padding all sequences to the maximum length results in considerable memory waste and fragmentation. To address this, vLLM~\cite{vllm} introduces PagedAttention, a KV cache management algorithm inspired by operating systems' virtual memory. Instead of allocating a continuous memory buffer for each sequence’s KV cache, PagedAttention partitions the cache into fixed-size blocks (or pages), each holding KV data for a set number of consecutive tokens (typically 16 to 64). A page block table records the physical address of each page, allowing the PagedAttention kernel to use indirect addressing to retrieve KV features. TensorRT-LLM~\cite{trtllm} and QServe~\cite{lin2024qserve} implement quantized page attention to reduce memory bandwidth usage during the decoding stage, resulting in further generation speedups.

\subsection{Motivation}
\label{sect:motivation}

\begin{figure}
    \centering
    \includegraphics[width=\linewidth]{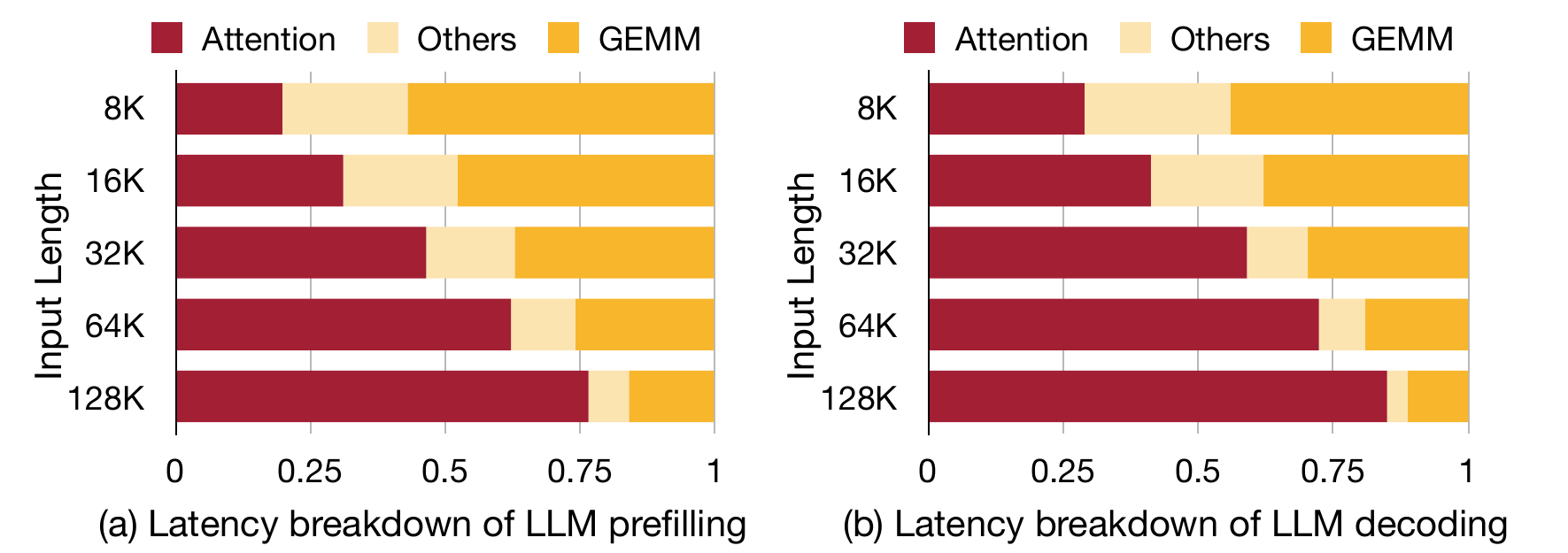}
    \caption{
    Latency breakdown of LLM inference during prefilling and decoding stages. Attention dominates both stages as sequence length increases, due to its quadratic complexity in prefilling and linear complexity in decoding. GEMM exhibits linear complexity in prefilling and constant complexity in decoding. Measurements obtained with Llama-3-8B on NVIDIA A100 GPU.
    } \label{fig:motivation:attention_ratio}
\end{figure}

Serving long-sequence LLMs is challenging due to the high cost of attention. Figure~\ref{fig:motivation:attention_ratio} profiles the latency breakdown of Llama-3-8B with a batch size of 1 across various sequence lengths on the A100 GPU. In both the prefilling and decoding stages, attention kernels account for at least 50\% of the runtime at sequence lengths over 64k, rising to 75\% at 128k. According to QServe~\cite{lin2024qserve}, the ratio of attention kernels in end-to-end runtime will increase as the batch size scale up. Therefore, in real-world serving scenarios, optimizing the attention becomes increasingly critical. 

\begin{figure}
    \centering
    \includegraphics[width=\linewidth]{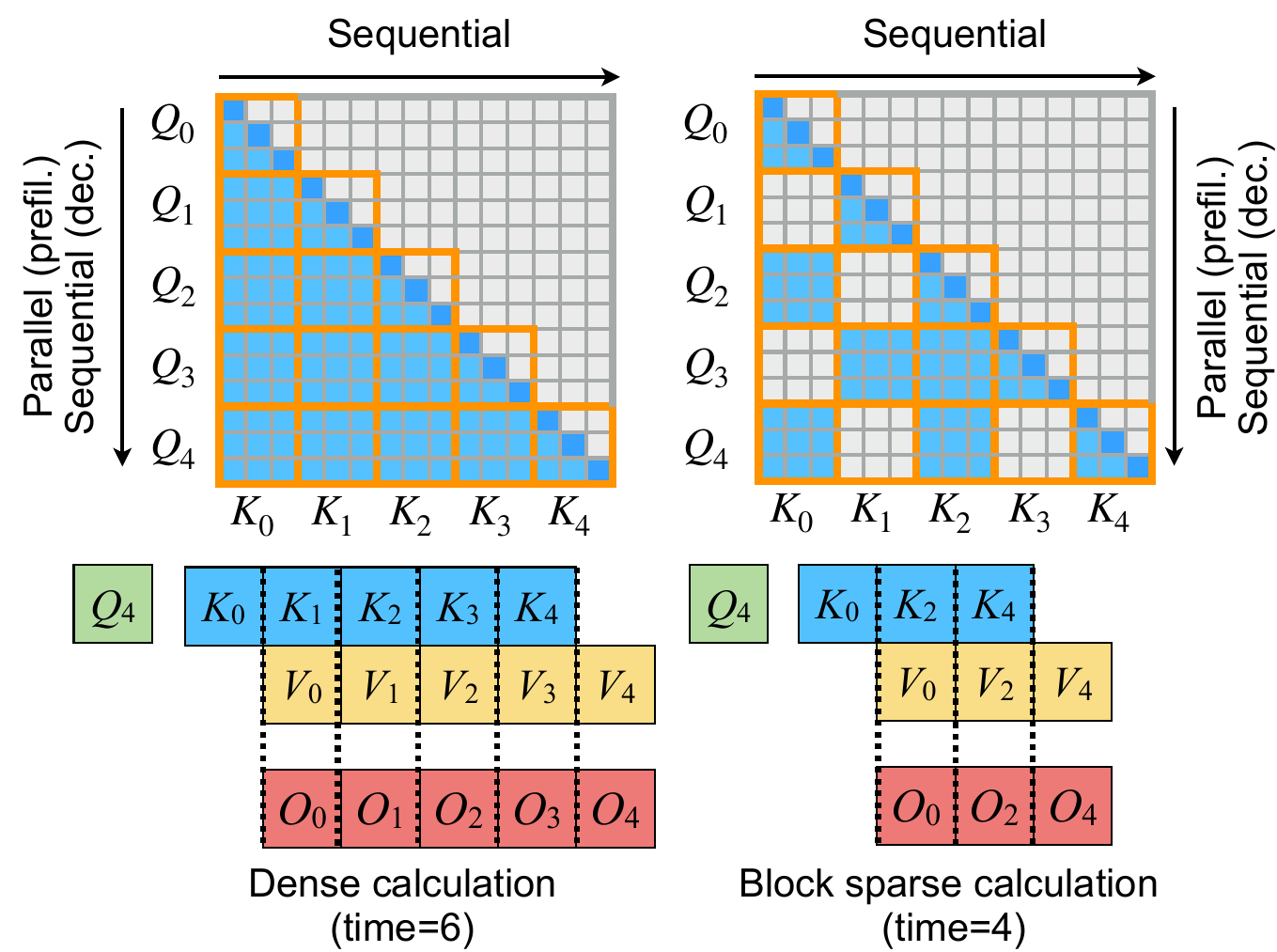}
    \caption{\textbf{Attention calculation on GPUs}: In both the decoding and prefilling stages, each query token iterates over all key and value tokens sequentially in a \textit{block-by-block} manner. Skipping KV blocks reduces the number of sequential iterations, directly accelerating attention.} %
    \label{fig:motivation:attention_flow}
    \vspace{-20pt}
\end{figure}

Accelerating attention in long-sequence LLMs requires a deep understanding of attention kernel implementation on GPUs, as illustrated in Figure~\ref{fig:motivation:attention_flow}. During the prefilling stage, the attention kernel is parallelized across batch size, attention heads, and query tokens, with query tokens set to 1 in the decoding stage. In both stages, the computation along the KV token dimension remains sequential. In each iteration, a block (depicted as a grid with an orange contour in Figure~\ref{fig:motivation:attention_flow}) is computed collaboratively by all threads in the current thread block. Although skipping certain computation within each block is possible, it yields minimal speedup. This is due to the lockstep execution of threads within a GPU warp, where faster threads are forced to wait for slower ones. 

That said, rather than focusing on sparsity within each iteration, a more effective way to accelerate attention is to \textbf{reduce the number of sequential iterations} along the KV token dimension. This approach leads to our unified \textit{block sparse attention} formulation, where attention computation is skipped in a blockwise manner. In this scheme, aside from the most recent KV block, each block is either fully computed or entirely skipped during the prefilling stage. During decoding, each sequence contains only one query token, reducing the dimensionality of each orange-contoured grid to 1$\times P$, where $P$ represents the page size (i.e., the number of KV tokens per page). We will detail \system's sparsity pattern selection in Section~\ref{sect:method}.

Additionally, because the decoding stage is memory-bound, KV cache quantization also contributes to speed improvements. Quantization is orthogonal to block sparsity, as it reduces the \textit{runtime of each iteration}, while sparsity reduces the \textit{number of iterations}. %

\begin{figure}[t]
    \centering
    \includegraphics[width=0.85\linewidth]{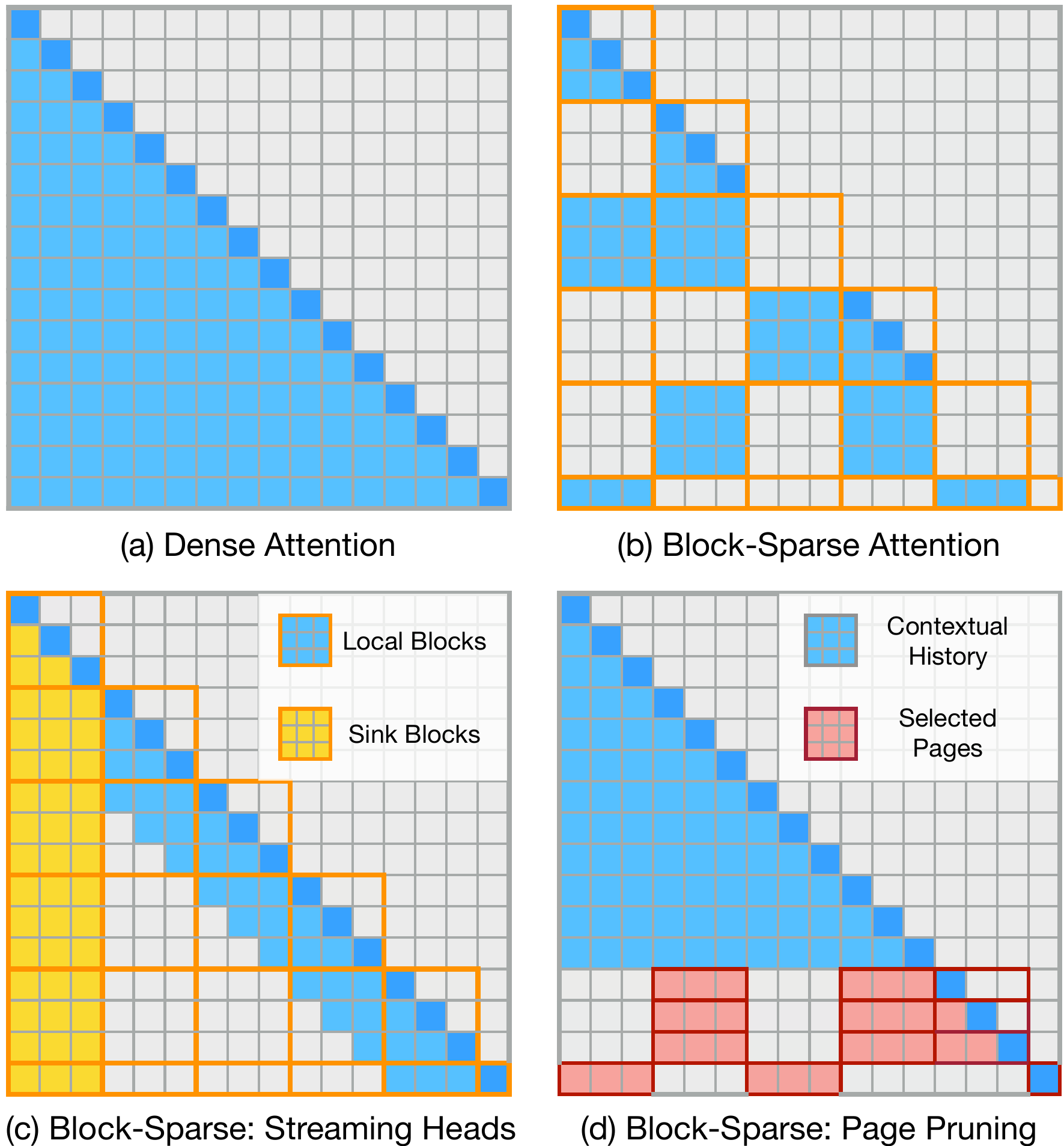}
    \caption{Unified block sparse attention pattern. \system integrates various sparsity patterns into a unified framework. 
    }
    \label{fig:method:blocksparse}
\end{figure}

\section{\system: Long-sequence Serving with Unified Sparse Attention}
\label{sect:method}

\begin{figure*}[t]
    \centering
    \includegraphics[width=\linewidth]{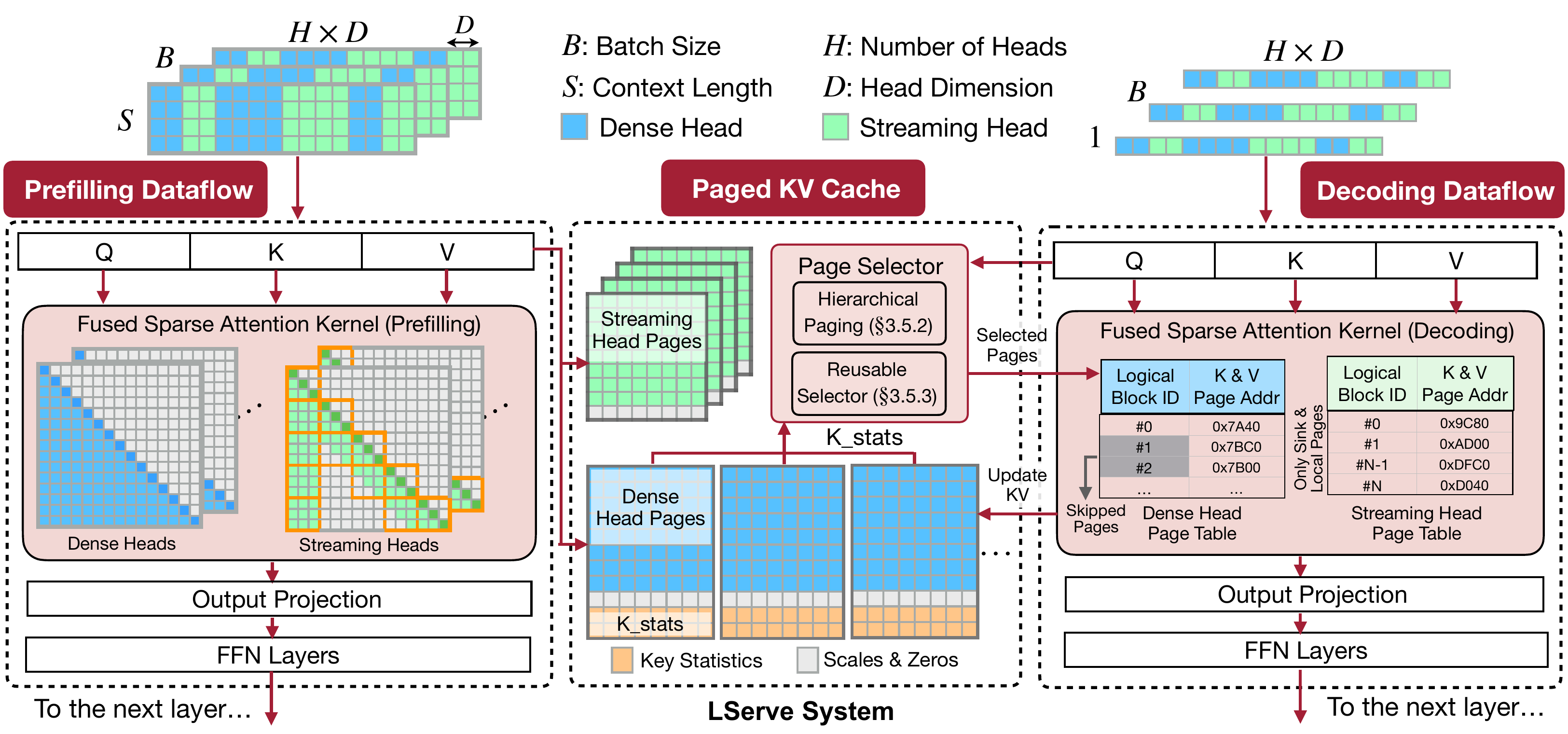}
    \caption{\system system overview. 
    In prefilling stage, \system processes both dense heads and streaming heads within a fused sparse attention kernel. Past Keys and Values are stored in two separate paging systems: one for streaming heads and the other for dense heads. In decoding stage, \system applies dynamic sparsity on dense heads with a page selection procedure. Only selected KV Pages will be loaded for the decoding stage attention.
    We omit normalization layers and residual connections in this figure for the sake of simplicity. }

    \label{fig:intro:overview}
\end{figure*}

We introduce \textbf{\system}, an efficient long-sequence LLM serving system featuring sparse attention. In \system, diverse sparse attention patterns are unified within a block-sparse formulation (Figure \ref{fig:method:blocksparse}), and are flexibly supported through fused CUDA kernels. \system also supports weight, activation and KV quantization, which significantly improves generation throughput at shorter context lengths. 

\subsection{Unified Block Sparse Attention}
\label{sect:method:blocksparse_overview}

As shown in Figure~\ref{fig:motivation:attention_flow}, skipping computations in the attention kernel by blockwise processing accelerates execution by shortening the sequential loop.
Building on this, we introduce a \textit{unified block sparse attention} pattern for both the prefilling and decoding stages: each thread block computes a $T_Q \times T_K$ tile (and $T_K \times T_V$) in parallel. Here, $T_Q > 1$ in the prefilling stage and $T_Q = 1$ in the decoding stage, with $T_K$ (or $T_V$) corresponding to the page size in PagedAttention~\cite{kwon2023efficient}.

We define \textit{block sparsity} in \system as follows: for each $T_Q \times T_K$ tile in the attention calculation, it is either fully skipped (Figure~\ref{fig:method:blocksparse}(b), light gray blocks) or retained as in standard causal attention (Figure~\ref{fig:method:blocksparse}(b), blue blocks). Given that each GPU streaming multiprocessor can execute only a limited number of thread blocks simultaneously, the attention kernel execution time can be approximated by the total count of $T_Q \times T_K$ (and $T_K \times T_V$) blocks. With a block sparsity of $r$, where $rN$ of the $N$ total blocks are empty, the theoretical speedup from block sparse attention is $1 / (1 - r)$. For example in Figure~\ref{fig:method:blocksparse}(b), 10 out of $N$=21 blocks are non-empty. Thus, the theoretical speedup ratio is 2.1$\times$. 

Figure~\ref{fig:method:blocksparse}(c)(d) shows two sparsity patterns used in \system. The first is streaming attention (Figure~\ref{fig:method:blocksparse}(c)), a specialized form of block-sparse attention where each token only attends to its immediate neighbors and initial tokens, known as attention sinks~\cite{xiao2023efficient}. Unlike dense attention, where computation for each row scales with the token index, streaming attention keeps the computation for each token \textit{constant}—in this case, only two local blocks and one sink block, as shown in Figure~\ref{fig:method:blocksparse}(c). This pattern is nearly cost-free in applications with extremely long contexts. Because streaming attention follows a fixed pattern, we designate which heads use it in \textit{offline}, and make it \textit{static} for different input sequences in both prefilling and decoding.

The second type of sparsity, illustrated in Figure~\ref{fig:method:blocksparse}(d), is page sparsity, which is specifically designed for the decoding stage where $T_Q=1$ applies to both skipped and selected pages. Unlike streaming attention, page sparsity in \system is \textit{dynamic}, allowing different query tokens to attend to different KV pages. As noted in Deja Vu~\cite{liu2023deja}, dynamic sparsity results in higher compression ratios than static sparsity. Our observations indicate that static sparsity offers up to a 2$\times$ efficiency gain, whereas dynamic sparsity bounds the decoding complexity to a \textit{constant}, with each query attending only to a fixed number of KV tokens.

\subsection{\system System Overview}
\vspace{2pt}

We present an overview of \system in Figure~\ref{fig:intro:overview}. Built on QServe, which natively supports quantized LLMs, \system enhances the baseline system by introducing sparsity into both prefilling and decoding dataflows. The \textit{two-way paged KV cache} serves as the bridge between these two stages. 

As discussed in Section~\ref{sect:method:blocksparse_overview}, we statically partition the attention heads of a pretrained LLM into two groups: dense heads and streaming heads. Unlike conventional LLM serving systems, which maintain a single KV cache, we utilize \textit{separate} KV caches for the dense and streaming heads. The KV cache for the streaming heads is organized similarly to the pages in QServe, with scaling factors and zero points stored immediately after the token features. Additionally, the KV cache for the dense heads includes \textit{key statistics} that facilitate critical page selection during the decoding stage.

In the prefilling stage, the key differences between \system and conventional dense-attention LLM serving systems are twofold: (1) we replace the dense attention kernel with our unified block sparse attention kernel, and (2) we write back quantized KV features using two distinct kernels. 

In the decoding stage, our system incorporates dynamic attention sparsity. Rather than developing an entirely new dynamic sparse attention kernel, we decompose the problem into two components: (1) dynamic \textit{page selection} and (2) a \textit{dense} attention kernel with \textit{shorter page tables}, where the shorter page tables are provided by the page selector. Notably, our page selector employs hierarchical paging and reusable page selection, enhancing both long-context accuracy and page selection efficiency.
\begin{figure}[t]
    \centering
    \includegraphics[width=\linewidth]{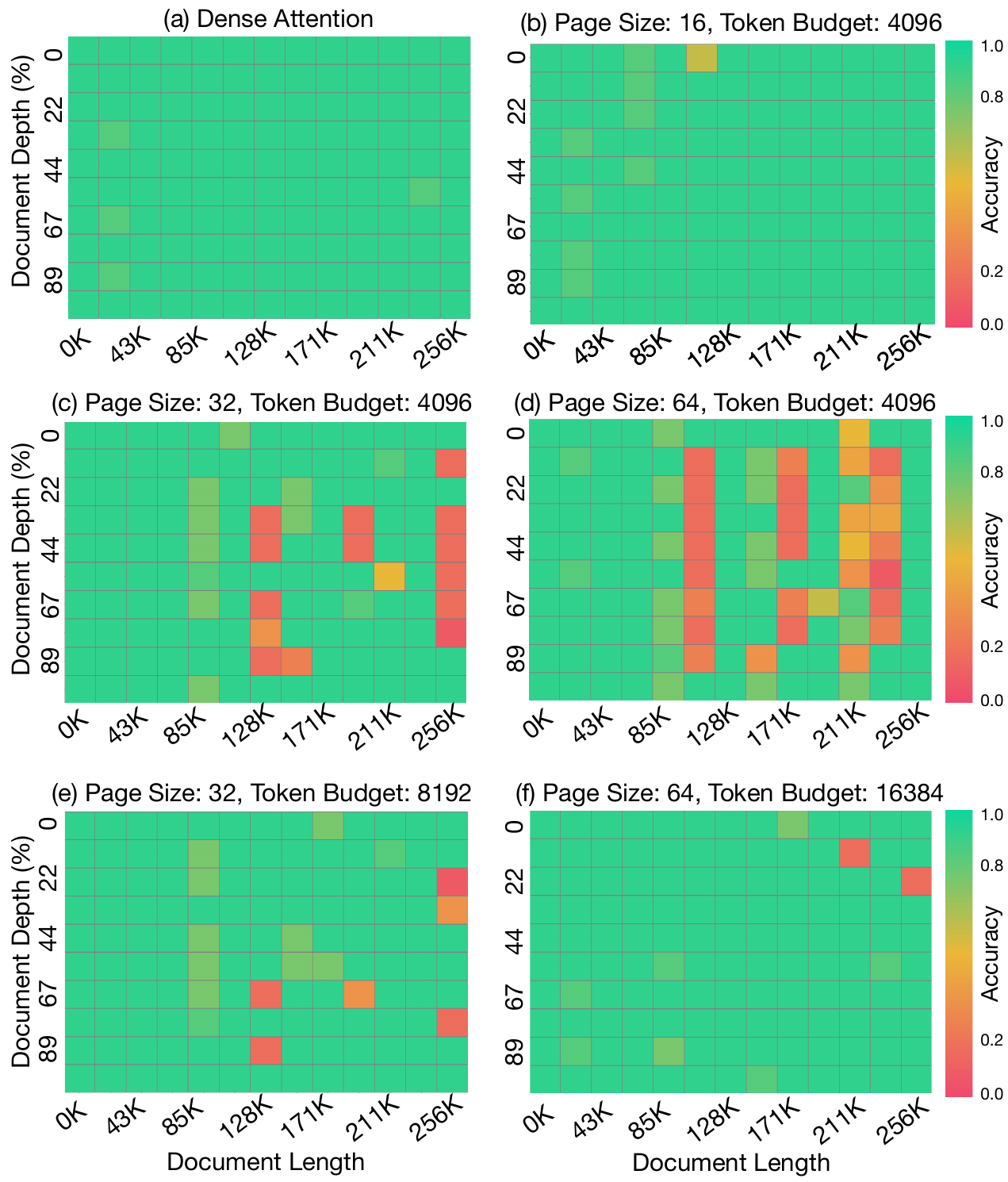}
    \caption{We evaluate the Llama-3-8B model with the Needle-in-a-Haystack (NIAH)~\cite{LLMTest_NeedleInAHaystack} benchmarks. The effectiveness of query-aware page selection algorithms (e.g., Quest~\cite{tang2024quest}) gets impaired when the KV page granularity grows (b,c,d). \textbf{Naively scaling up the page sizes will lead to significant performance loss} even if we linearly increase the number of selected pages (token budget) (e,f). }
    \label{fig:ana:naive-larger-page}
\end{figure}

\subsection{Prefilling Stage: Sparsity Determination}

We adopt the approach from DuoAttention~\cite{xiao2024duoattention} to classify each attention head as either a retrieval head or a streaming head. Using DuoAttention's optimization-based identification method, we obtain a gating value $\alpha \in [0, 1]$ for each head, where values closer to 1 signify a retrieval head, and values closer to 0 indicate a streaming head. To classify a head as a retrieval head, we compare $\alpha$ to a threshold $\tau$, determined by a sparsity quantile. For instance, with a target sparsity of 50$\%$ across attention heads, $\tau$ equals the median of all gate values, thereby designating half of the heads as retrieval heads.

\subsection{Prefilling Stage: Kernel Implementation}

To effectively translate sparsity into performance gains, it is essential to avoid iterating over a complete sequential loop and relying on conditional statements to determine data loading and computation requirements. This method is inefficient for GPU computation patterns, which thrive on minimizing branching within loops. Instead, we should focus on iterating only over the necessary blocks by accurately calculating offsets to load data and assess whether a block should be processed.

To facilitate this, we introduce an iterator-based abstraction that standardizes indexing operations. This allows us to loop exclusively over the blocks requiring computation, with data offsets easily computed using 
$\text{offset} = \text{iter}(i + 1) - \text{iter}(i)$. This abstraction efficiently skips unnecessary blocks with minimal overhead and necessitates few changes to the kernel function, thus enhancing maintainability. Take the streaming heads as an example, the iterators are determined outside the attention kernel since streaming heads are configured offline and the attention pattern is fixed. Once the attention on sink tokens is complete, the iterator automatically updates the memory pointer to the first local token in the KV cache with minimal overhead. Additionally, our iterator-based formulation unifies the more general block sparse pattern (see Figure~\ref{fig:method:blocksparse}).

\subsection{Decoding Stage: Sparsity Determination}
\label{sect:method:decoding_sparsity_determination}

To further enhance the long-context LLM decoding throughput, we introduce dynamic sparsity upon the input-agnostic static sparsity in Sec.~\ref{sect:method:blocksparse_overview}. 

\subsubsection{Challenge: the Page Size Dilemma}

\begin{table}[t]
\centering
\caption{Page size significantly impacts the LLM serving system's efficiency: Larger page size is more hardware-friendly as it improves contiguity of memory layout and the GPU bandwidth utilization during attention computation. For example, simply shrinking the page size in QServe~\cite{lin2024qserve} leads to prominent slow-down of the end-to-end system. We evaluate the per-step decoding latency (ms / step) of QServe on a single A100 GPU for demonstration. We use Llama3-8B model architecture, with the batch size of 32.
}
\vspace{5pt}
\footnotesize
\scalebox{1.0}{
\begin{tabular}{ccccc}
\toprule
\multirow{2.5}{*}{Seq\_len} 
& \multicolumn{4}{c}{Page Size} \\ 
\cmidrule{2-5} & 16 & 32 & \textbf{64} & 128 \\
\midrule
512 & 11.0 ms & 10.7 ms & \textbf{10.5 ms} & 10.5 ms\\
1024 & 13.8 ms & 13.0 ms & \textbf{12.7 ms} & 12.7 ms\\
2048 & 22.1 ms & 20.1 ms & \textbf{18.3 ms} & 18.2 ms\\
4096 & 35.7 ms & 31.6 ms & \textbf{28.1 ms} & 28.1 ms\\
8192 & \textcolor{red}{\textbf{77.1 ms}} & \textcolor{red}{\textbf{63.0 ms}} & \textbf{51.0 ms} & 50.6 ms\\
\midrule
Max Slowdown & 1.52$\times$ & 1.25$\times$ & 1.01$\times$ &  1.00$\times$ \\
\bottomrule
\end{tabular}
}
\label{tab:ana:page_size_speed}
\end{table}

In the decoding stage, the attention operation is memory-bound, so state-of-the-art systems typically implement KV cache quantization to reduce device memory usage and enhance throughput. However, this quantization introduces challenges for further optimization. Specifically, reducing the bit-width of KV tokens necessitates larger page sizes to maintain GPU memory bandwidth utilization. Failure to do so can lead to significant throughput loss (Table~\ref{tab:ana:page_size_speed}). Yet, larger KV page sizes complicate the sparsification process; for example, Quest~\cite{tang2024quest}, which estimates token criticality using page-wise statistics, fails when page sizes increase (Figure~\ref{fig:ana:naive-larger-page}). This observation poses challenges to balance between accuracy and efficiency.

\begin{figure}[t]
    \centering
    \includegraphics[width=\linewidth]{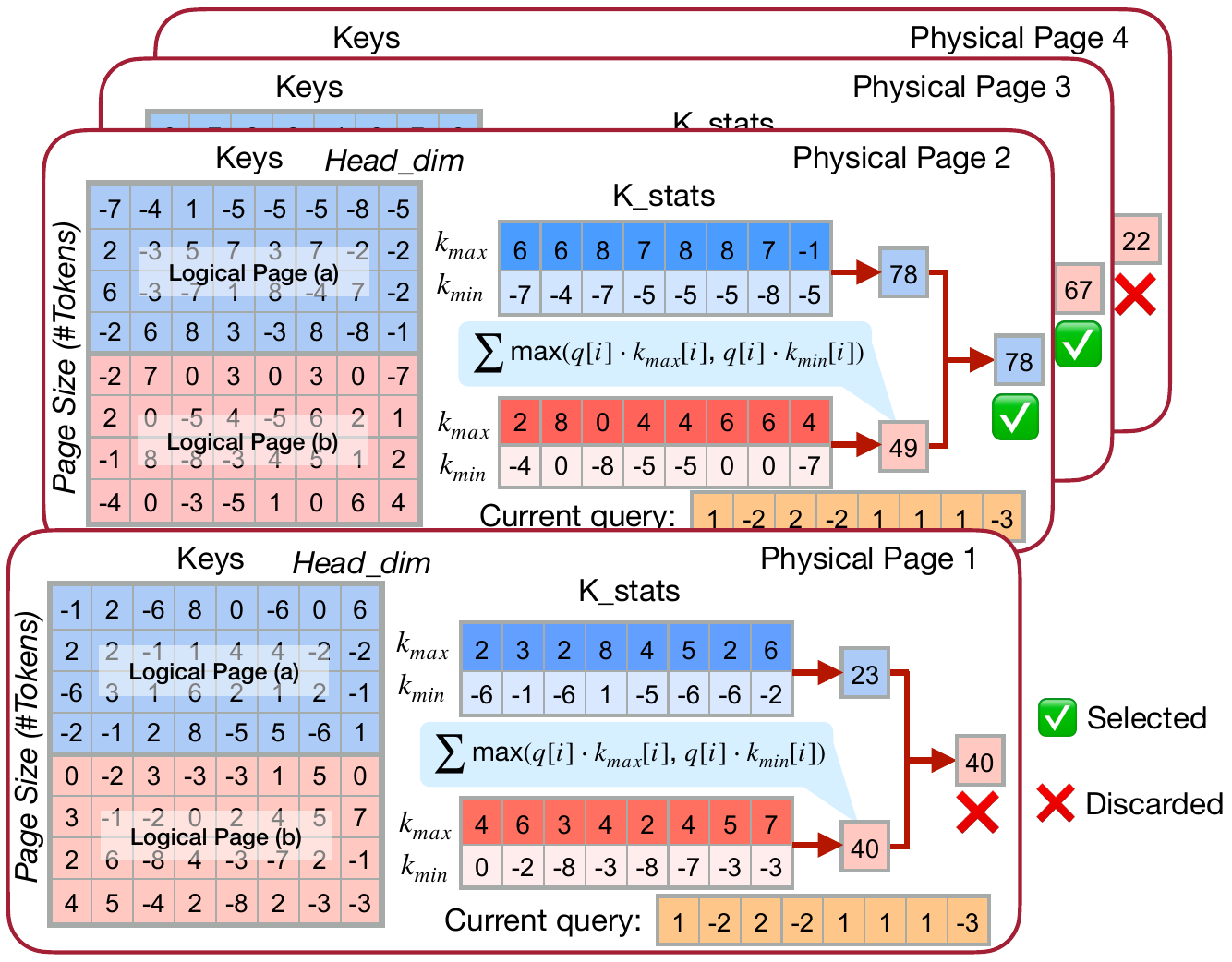}
    \caption{Hierarchical Paging in \system system. We assume the each \textit{physical page} contains $N_p = 8$ tokens and each \textit{logical page} has $N_l = 4$ tokens. The $k_{max}$ and $k_{min}$ vectors are concatenated to the end of each \textit{physical page}, and are pre-computed during the context stage and previous decoding steps. The importance of each \textit{physical page} is decided by the max of the importance scores of the \textit{logical pages} it contains. We omitted KV quantization in this figure for the sake of simplicity. %
    }
\label{fig:method:page_selector}
\end{figure}

\begin{figure}[t]
    \centering
    \includegraphics[width=\linewidth]{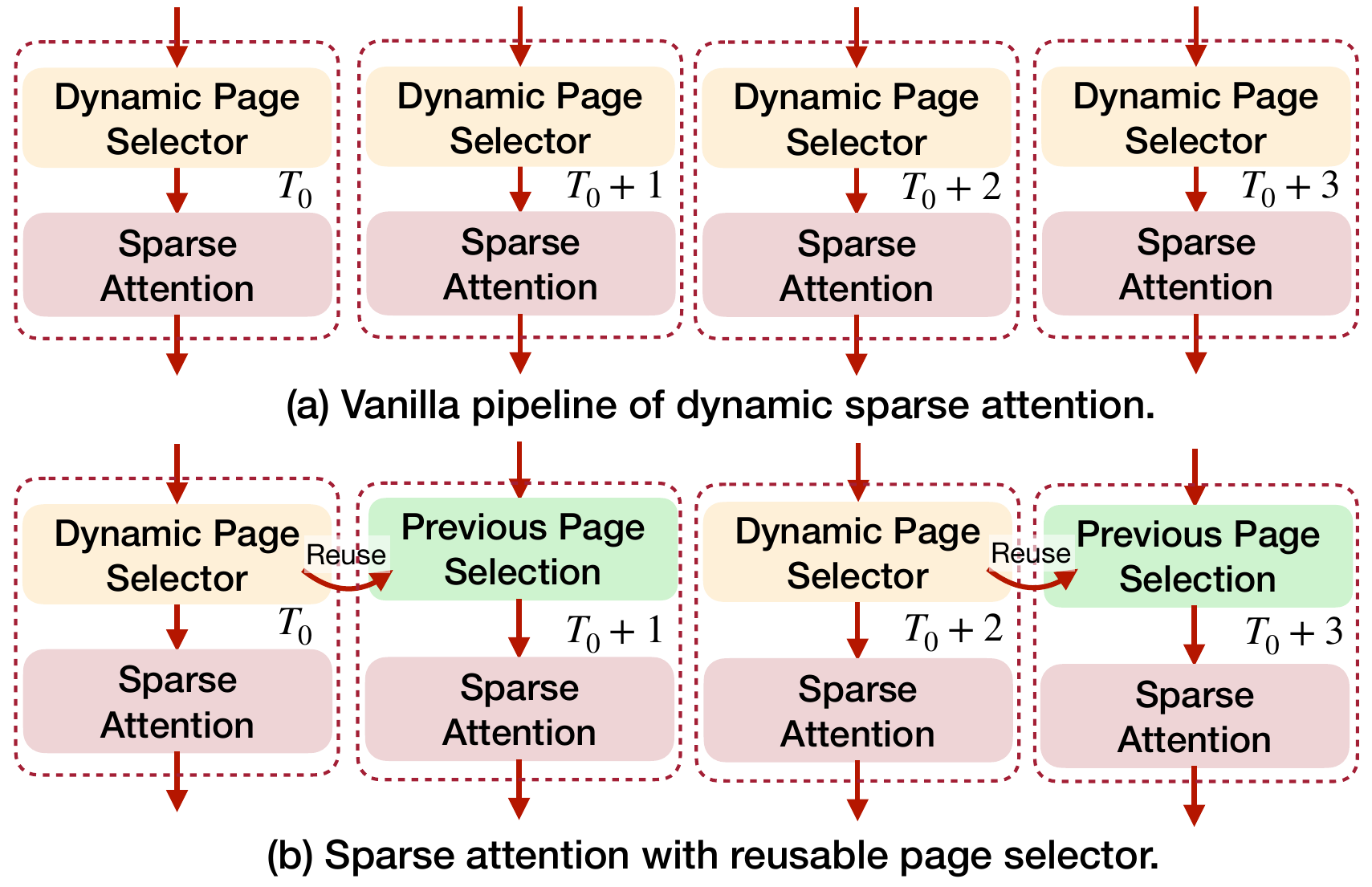}
    \caption{We introduce Reusable Page Selector in \system, which utilize the similarity of queries of consecutive tokens to cut down the selector overhead. The chunk size of reusable selector is set to 2 in this figure for the demonstration purpose.
    }
\label{fig:method:reusable_selector}
\end{figure}

\subsubsection{Hierarchical Paging: Mitigating the accuracy-efficiency tradeoff} 
\label{sect:method:decoding:hierarchical_page}

We observe that the failure of query-aware KV cache selection paradigm (Figure~\ref{fig:ana:naive-larger-page}) is not due to the coarser granularity of sparse attention (i.e., larger page size). Rather, the underlying cause lies in that page-wise statistical indicators become homogenized and less representative especially when there are excessive tokens within a single page.
To address this issue, we design a simple-yet-effective hierarchical paging system that introduces an abstract layer of virtual \textit{logical page} for estimating token criticality, while preserving the original memory layout of KV cache in (\textit{physical pages}).
As illustrated in Figure~\ref{fig:method:page_selector}, our hierarchical paging groups $N_L$ tokens into a logical page and $N_P$ tokens into a physical page ($N_P = g\cdot N_L, g\in \mathbb{Z}$), that is, a physical page contains $g$ logical pages. 
Tokens within the same logical page will collectively contribute to the same criticality estimator. In \system, we utilize the channel-wise minimum and maximum values of keys in the same logical page as its representative vectors, which has been proven to be an effective metric~\cite{tang2024quest} for page importance estimation with a moderate page size ($\leq 16$).
The current query will attend to representative vectors of each logical page to calculate the corresponding importance score as follows:

\vspace{-14pt}
\begin{equation}
\small
S^{j} = \sum_i^D \max{\left(q[i]*k^j_{max}[i], \hspace{3pt} q[i]*k^j_{min}[i] \right)}
\label{eqn:method:estimator}
\end{equation}
\vspace{-14pt}

where $S$ is the importance score of logical page, $j\in\{a,b,...\}$ is the index of logical page, $i$ is the channel index, and $D$ refers to the head dimension.

The importance of each physical page is determined by the max-reduction over the importance scores of its corresponding logical pages. Finally, \system selects the top-K physical pages (based on the predefined token budget) with highest importance scores as the input of sparse attention kernel.

\subsubsection{Reducing sparse attention overheads with locality}
\label{sect:method:decoding:locality}

One remaining question is: as physical page size increases, will the hierarchical paging require a higher token budget for sparse attention to retain accuracy?

Given a generation step, assume the most important history tokens are distributed in a logical page set $\mathcal{P}=\{P(i,j)\}$, where $i\in\{1,2,...\}, j\in\{a,b,,...\}$ are the physical and logical index of a page accordingly. If these important tokens are randomly and sparsely distributed in the context, chances are that all logical pages in $\mathcal{P}$ are scattered in different physical pages, that is, for any $P_1, P_2 \in \mathcal{P}$, $i_1 \neq i_2$. In this case, all $|\mathcal{P}|$ physical pages ($|\mathcal{P}|\cdot N_P$ tokens) are selected to avoid losing important information. However, the \naive paging only needs to keep $|\mathcal{P}|\cdot N_L$ tokens since it directly shrinks page sizes to a smaller granularity (e.g., $N_L$). Consequently, our hierarchical paging may suffer from a decrease in attention sparsity by $N_P/N_L$.

Fortunately, the semantic continuity of natural language endows the attention operation with intrinsic locality, allowing \system to maintain a consistent sparse attention token budget for larger physical page sizes.
During the decoding stage, the coherence of contextual tokens makes the current query token incline to attend to consecutive pages in the KV cache. As a result, logical pages with highest importance scores tend to cluster within similar physical pages. This kind of \textit{spatial locality} effectively alleviates the need for a increased token budget, thereby reducing the overhead caused by the contradiction between quantization and sparse attention. Experimental results in Figure~\ref{fig:ana:our_larger_page} further affirm that our hierarchical paging well preserves the model accuracy even with the same token budget as the vanilla page selector with smaller page sizes.

Moreover, the attention mechanism in decoding stage also exhibits the \textit{temporal locality}: adjacent query tokens also heavily attend to similar historical pages. And there is no need for queries at consecutive decoding steps to select salient pages independently. Instead, the page selection decision can be shared across queries, aligning with the block-sparse attention formulation illustrated in Figure~\ref{fig:method:blocksparse}(d).

To this end, we present \textit{Reusable Page Selection} in \system. As in Figure~\ref{fig:method:reusable_selector}, we activate the page selector only at the very beginning of pre-defined chunks. For the consecutive tokens within the same chunk, we reuse the page selection results from the first token of the chunk.
Utilizing the temporal sparsity of attention, reusable page selection substantially improves the long-context generation speed by a great margin without sacrificing accuracy. As demonstrated in Figure~\ref{fig:ana:selector_overhead}, even though dynamic sparse attention effectively restrict the complexity of decoding attention, the latency of page selector increases linearly with regard to the sequence length. When the number of history tokens surpasses 64K, the \naive page selector becomes the bottleneck to system efficiency, whereas our reusable page selector significantly alleviates this problem.

\subsection{Decoding Stage: Kernel Implementation}

During the decoding stage, attention heads are processed in parallel on GPU, enabling different sparsity patterns to be applied independently on each head. This flexibility enables some heads to operate with page-level sparsity while others follow the streaming computation pattern. 

To leverage this, we employ a two-level indexing hierarchy to unify the operations for streaming heads and dense heads with dynamic sparsity. Specifically, the low-level (physical) index corresponds to the iteration step of current GPU thread, which executes in a consecutive manner as in dense attention, while logical index denotes the actual position of the target token within the entire KV cache. For each dense head, the page selector provides an index table to map physical index to logical index. Streaming heads are treated as dynamic sparse heads with index table only containing the sink and local pages.

\section{Evaluation}
\label{sect:results}

\subsection{Evaluation Setup}

\textbf{Implementation}. We implement \system in CUDA and PTX assembly on the basis of QServe~\cite{lin2024qserve} and TensorRT-LLM~\cite{trtllm} system. The specialized CUDA kernels are compiled into PyTorch extensions for better flexibility and compatibility with the purely PyTorch-based serving interface.

\textbf{Testbed}. 
Our primary experiments are conducted on a server equipped with 8 NVIDIA A100 80GB GPUs, 2 AMD EPYC 7763 CPUs (128 cores), and 2TB of memory. Unless explicitly stated, all experiments utilize the A100 GPUs. Additionally, we perform some evaluations on a cloud instance with a single NVIDIA L40S 48GB GPU to assess system performance across different GPU architectures. All evaluations use PyTorch 2.5.0 with CUDA 12.4 and cuDNN 9.2.0.
\begin{table}[t]
\centering
\small
\caption{\textbf{Accuracy evaluation on LongBench}~\cite{bai2023longbench}. We compare our method with vanilla dense attention on 2 models and 10 different benchmarks.}
\scalebox{1.0}
{
\begin{tabular}{ccccc}
\toprule
Model & \multicolumn{2}{c}{Llama-3-8B} & \multicolumn{2}{c}{Llama-2-7B} \\
\midrule
Benchmark & Dense & \system & Dense & \system \\
\midrule
2WikiMQA  & 30.3 & 31.6 & 35.4 & 35.1 \\ 
DuReader  & 30.3 & 30.8 & 25.4 & 24.7 \\ 
HotpotQA  & 41.7 & 42.7 & 47.4 & 49.6 \\ 
MultiNews & 27.7 & 27.7 & 26.6 & 26.6 \\ 
Qasper    & 31.7 & 29.3 & 32.6 & 29.5 \\ 
QMSum     & 23.8 & 24.0 & 21.0 & 21.3 \\  
SamSum   & 41.2 & 39.3 & 41.8 & 41.5 \\ 
TriviaQA  & 84.9 & 83.7 & 86.2 & 86.5 \\ \midrule
\textbf{Average}   & \textbf{38.9} & \textbf{38.6} & \textbf{39.5} & \textbf{39.4}
\\
\bottomrule
\end{tabular}
}
\label{tab:results:long_bench}
\vspace{3pt}
\end{table}

\begin{figure}[t]
    \centering
    \includegraphics[width=\linewidth]{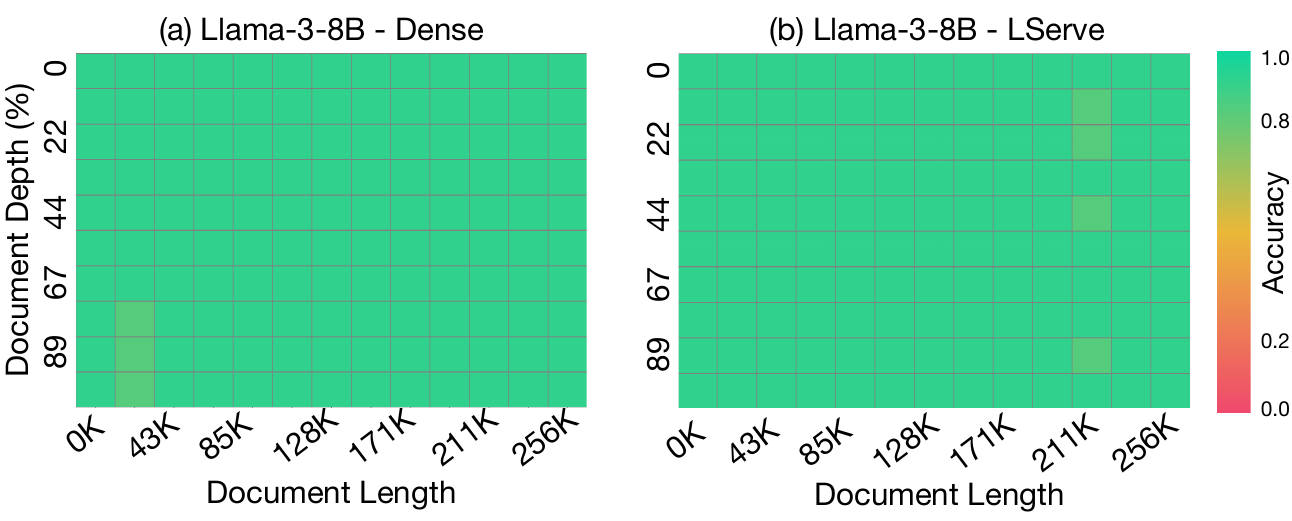}
    \caption{\textbf{Accuracy evaluation on Needle-in-a-Haystack}. 
    }%
    \label{fig:evaluation:main_niah}
    \vspace{10pt}
\end{figure}

\begin{table}[t]
\centering
\caption{\textbf{Accuracy evaluation on RULER}~\cite{nvidia_ruler}. We evaluate the accuracy of Llama-3-8B on RULER benchmarks, including challenging tasks such as multi-hop tracing and aggregation to test behaviors beyond searching from context. LServe-$N$ denotes that the token budget for dynamic sparsity is $N$. Note that for long-context inputs, latency is not dominated by attention alone in \system, with page selector and GEMM also contributing to it. Experiments reveal that LServe-8192 is only up to 6\% slower than LServe-4096 when the sequence length exceeds 128K.}
\scalebox{0.9}
{
\begin{tabular}{ccccccc}
\toprule
Llama-3-8B  & 32K  & 64K  & 128K & 160K	& 192K & 256K \\ \midrule
Dense       & 90.5 & 86.8 & 83.8 & 79.3 & 79.6 & 79.4 \\ \midrule
LServe-4096 & 91.0 & 85.6 & 81.0 & 79.0 & 76.1 & 75.7 \\ \midrule
LServe-8192 & 91.8 & 86.1 & 81.7 & 81.2 & 79.7 & 79.1 \\
\bottomrule
\end{tabular}
}
\label{tab:results:main_ruler}
\vspace{5pt}
\end{table}

\begin{table}[t]
\centering
\small
\caption{\textbf{Accuracy evaluation on AIME}~\cite{AIME} \textbf{and MATH500}~\cite{hendrycks2021measuring} DS-R1-Llama-8B refers to DeepSeek-R1-Distill-Llama-8B.~\cite{deepseekai2025deepseekr1incentivizingreasoningcapability}}

\scalebox{1.0}
{
\begin{tabular}{ccc}
\toprule
Model & \multicolumn{2}{c}{DS-R1-Llama-8B} \\
\midrule
Benchmark & Dense & \system  \\
\midrule
AIME@2024  & 43.3 & 43.3 \\ 
MATH500   & 84.2 & 85.4  \\  \midrule
\textbf{Average}   & \textbf{63.8} & \textbf{64.4}
\\
\bottomrule
\end{tabular}
}
\label{tab:results:reasoning}
\vspace{8pt}
\end{table}

\textbf{Models}. 
To comprehensively assess system performance across various LLM architectures, we utilize the widely adopted GQA-based model Llama-3-8B \cite{dubey2024llama}, the MHA-based model Llama-2-7B \cite{touvron2023llama2}, and the smaller-scale model Minitron-4B \cite{Minitron}. Additionally, to support long-context inference, we employ the context-extended Llama-3-8B version Gradient \cite{gradientlongcontextllama3}.

\textbf{Metrics}.
Our primary focus is on serving throughput. For the prefilling stage, we use \emph{time-to-first-token} (TTFT) as a key metric, while for the decoding stage, we emphasize minimizing the \emph{per-token generation latency}.

\textbf{Baselines}. 
We consider the following systems as baselines, using their latest versions\footnote{vLLM 0.6.3} to ensure a fair comparison. We activated W8A8 precision for baselines if available.

\begin{itemize}[leftmargin=*, itemsep=-3pt, topsep=-5pt, ]
  \item \emph{vLLM}~\cite{vllm}, one of the most popular LLM serving systems featuring PagedAttention.
  
  \item \emph{QServe}~\cite{lin2024qserve}, efficient LLM serving system featuring W4A8KV4 quantization.
  
  \item \emph{MInference} \cite{jiang2024minference}, the state-of-the-art long-context prefilling stage acceleration system.
  
  \item \emph{DuoAttention} \cite{xiao2024duoattention}, a strong long-sequence LLM inference framework with static sparse attention.
  
\end{itemize}

Additionally, we compare our approach with the state-of-the-art long-context decoding stage acceleration system, \emph{Quest} \cite{tang2024quest}. Since Quest only supports MHA models, we conduct and discuss this comparison in Table~\ref{tab:results:e2e_quest}.

\subsection{End-to-end Accuracy}
\label{sect:results:acc_eval}

We evaluate the accuracy of our hybrid block-sparse mechanism with LongBench~\cite{bai2023longbench} tasks, the Needle-in-a-Haystack (NIAH)~\cite{LLMTest_NeedleInAHaystack} pressure tests, as well as the challenging RULER~\cite{nvidia_ruler} benchmarks.  Table~\ref{tab:results:long_bench} compares the LongBench accuracy between \system and dense baseline. Results show that \system well preserves the performance of two models across different test sets. Figure~\ref{fig:evaluation:main_niah} showcases the NIAH evaluation results of our system, where \system also achieves the same level of accuracy compared to the dense baseline. In Table~\ref{tab:results:main_ruler}, we test \system with RULER benchmarks. Unless otherwise specified, we convert half of the attention heads into streaming heads and keep token budget for dynamic sparsity to 4096 for the benchmarks.

Additionally, we also evaluate \system 
on complex reasoning tasks such as AIME2024~\cite{AIME} and MATH500~\cite{hendrycks2021measuring}. As shown in Table~\ref{tab:results:reasoning}, \system maintains accuracy comparable to its dense counterparts on these complex reasoning tasks.

\begin{figure*}[t]
    \centering
    \includegraphics[width=\linewidth]{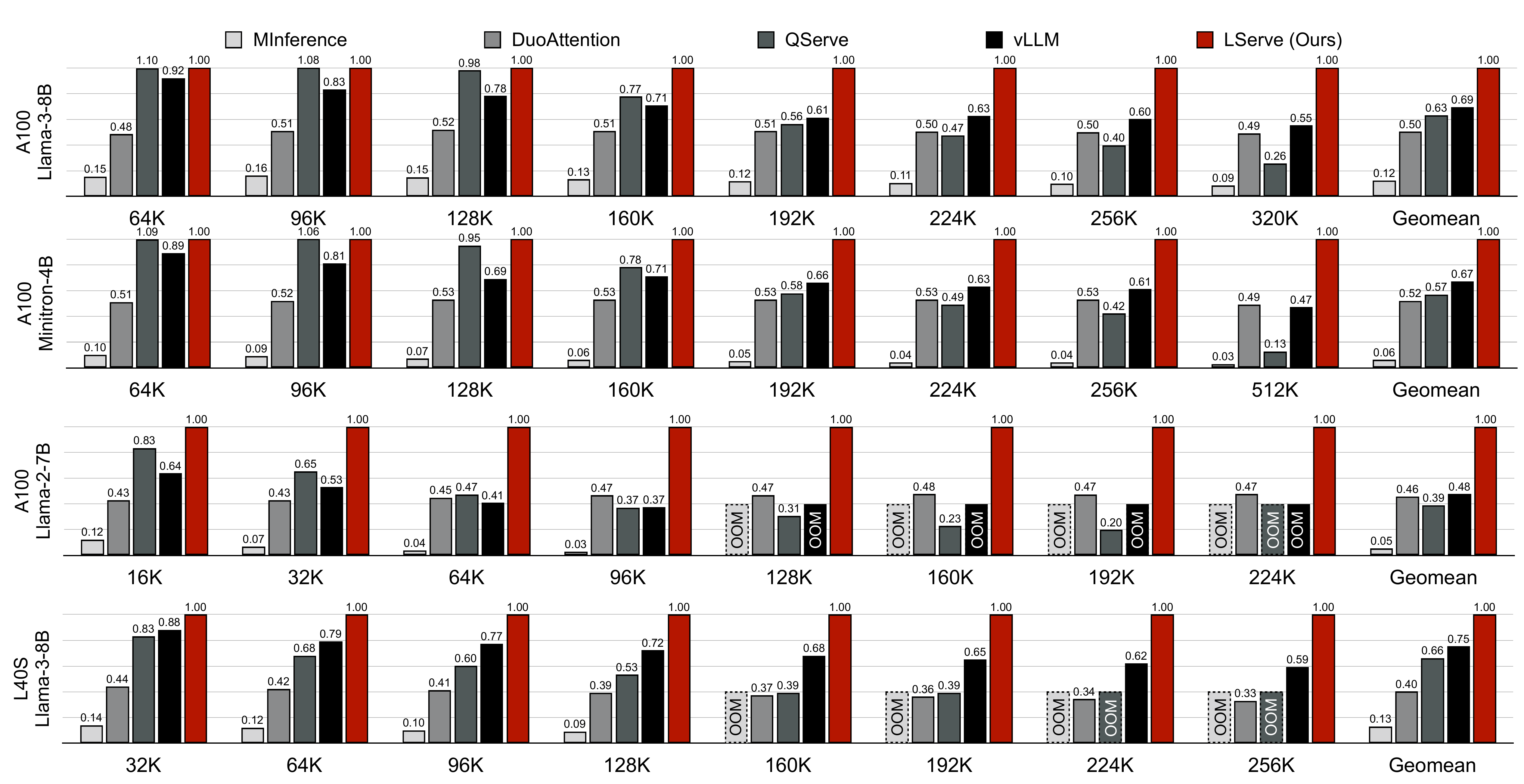}
    \caption{\textbf{Decoding Speed Evaluation}. The y-axis indicates the relative throughput of each system, normalized by the speed of \system. Note that MInference exhibits limited decoding performance due to its unoptimized decoding stage with \textbf{dense attention}, but when integrated into vLLM, it can achieve throughput comparable to that of vLLM. 
    }
    \label{fig:evaluation:main_speed}
    \vspace{-5pt}
\end{figure*}

\begin{figure}[t]
    \centering
    \includegraphics[width=\linewidth]{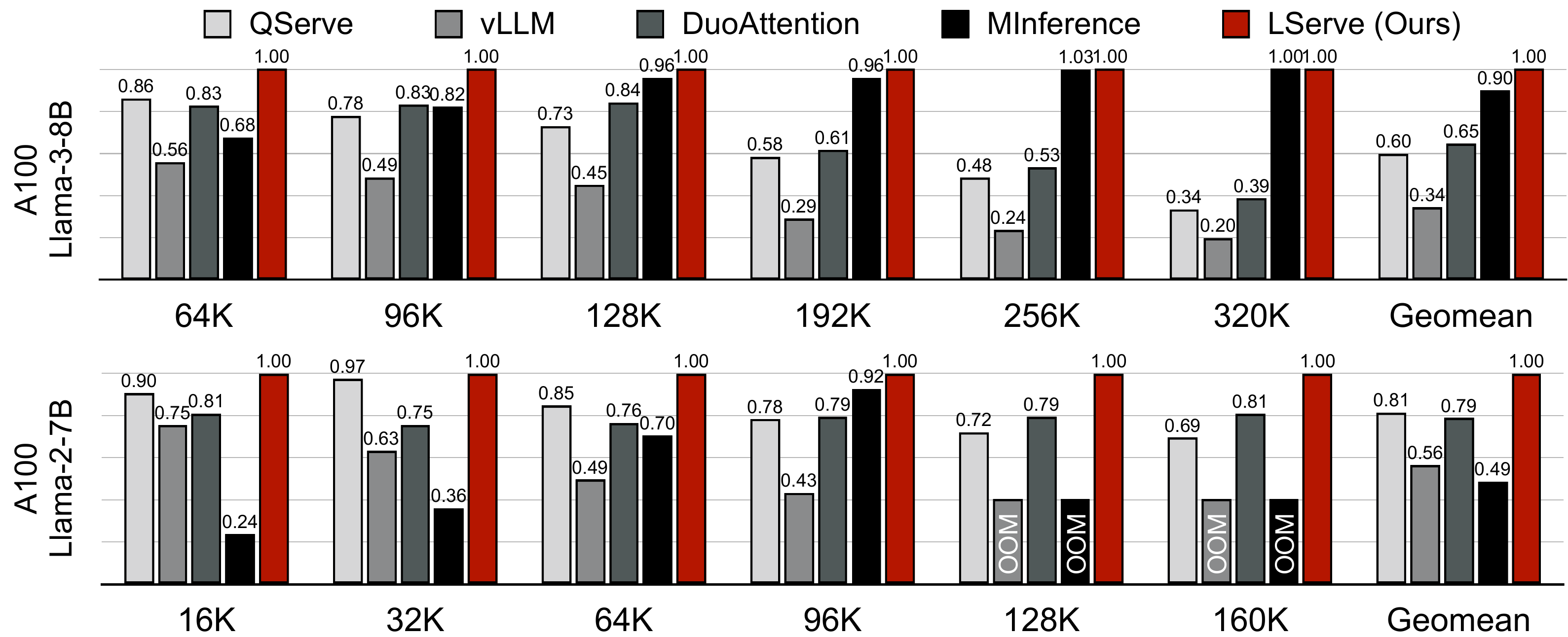}
    \caption{\textbf{Prefilling Speed Evaluation}. Performance comparison of long-sequence prefilling across different serving frameworks, normalized to \system's speed.
    }%
    \label{fig:evaluation:main_speed_prefill}
\end{figure}

\subsection{End-to-end Efficiency}

\textbf{Decoding Efficiency}. Figure~\ref{fig:evaluation:main_speed} presents the efficiency benchmarking results for the decoding stage. We use the same sparsity configurations as in \sect{sect:results:acc_eval}. Compared with the state-of-the-art serving systems, \system demonstrates significant and consistent efficiency improvements across different GPU platforms and model architectures. On Llama-3-8B and Minitron-4B, \system achieves 1.5$\times$ average speedup over vLLM. For MHA-based model Llama-2-7B, \system runs more than 2.0$\times$ faster than baselines on average. Additionally, we demonstrate that \system also functions well on other GPU devices such as L40S with Ada Lovelace Architecture. \system achieves up to 1.7$\times$ speedup over vLLM.

\textbf{Prefilling Efficiency}. In Figure~\ref{fig:evaluation:main_speed_prefill}, we compare the prefilling speed of \system against 4 baselines on Llama-3-8B and Llama-2-7B. \system maintains superior prefilling throughput across different sequence lengths. For instance, on Llama-2-7B, \system achieves an average of 1.8$\times$ higher prefilling throughput over vLLM. \system is also compatible with the prefilling dynamic sparsity in MInference, which we activated after 128K sequence length.

\begin{table}[t]
\centering
\caption{\system achieves lower latency over Quest~\cite{tang2024quest} system in both prefilling stage and decoding stage. We benchmark the two systems on Llama-2-7B model, since Quest does not support GQA~\cite{ainslie2023gqa} architecture. }
\scalebox{0.8}
{
\begin{tabular}{cccccccccc}
\toprule
\multirow{2.5}{*}{Stage}      & \multirow{2.5}{*}{System} & \multicolumn{5}{c}{Sequence Length} \\ \cmidrule{3-7} 
 & & 4K & 8K & 16K & 32K & 64K \\
\midrule
\multirow{4}{*}{\makecell[l]{Prefilling \\ Latency (s)}} & Quest & 0.51    & 0.82   & 1.62  & 3.61    & OOM      \\ \cmidrule{2-7}
& \system  & 0.24     & 0.49     & 1.08    & 2.32    & 5.27    
 \\ \cmidrule{2-7}
     
& Speedup & 2.1 $\times$      & 1.7$\times$       & 1.5$\times$     & 1.6$\times$     & /        \\ \midrule

\multirow{4}{*}{\makecell[l]{Decoding \\ Latency (ms)}}   & Quest & 13.13   & 13.58   & 14.08  & 14.86  & OOM      \\ \cmidrule{2-7}
& \system  & 10.02   & 10.29    & 10.22  & 10.24  & 11.54    \\ \cmidrule{2-7}
& Speedup    & 1.3$\times$      & 1.3$\times$       & 1.4$\times$     & 1.5$\times$     & /        \\ 
\bottomrule
\end{tabular}
}
\label{tab:results:e2e_quest}
\end{table}

\vspace{-10pt}
\subsection{End-to-End Comparison with Quest}

We also compares our system against Quest~\cite{tang2024quest} in Table~\ref{tab:results:e2e_quest}. Across different sequence lengths, \system consistently outperforms Quest in both prefilling (1.6-2.1$\times$ speedup) and decoding stages (1.3-1.5$\times$ speedup).

\section{Analysis}
\label{sect:analysis}

\begin{figure}[t]
    \centering
    \includegraphics[width=0.86\linewidth]{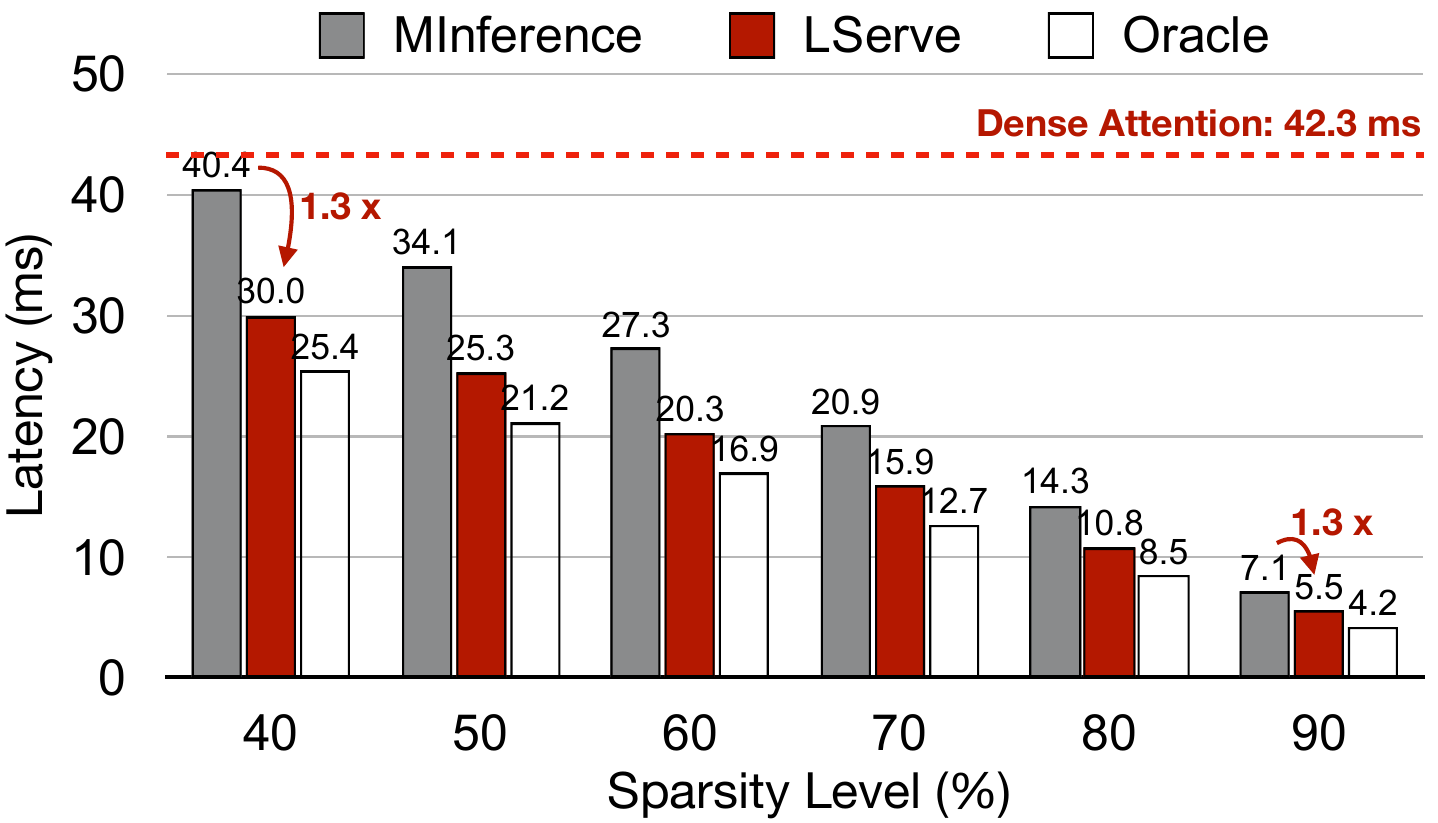}
    \caption{Prefilling Stage Attention Kernel Evaluation.} 
        
    \label{fig:ana:prefilling_attention}
\end{figure}

\begin{figure}[t]
    \centering
    \includegraphics[width=\linewidth]{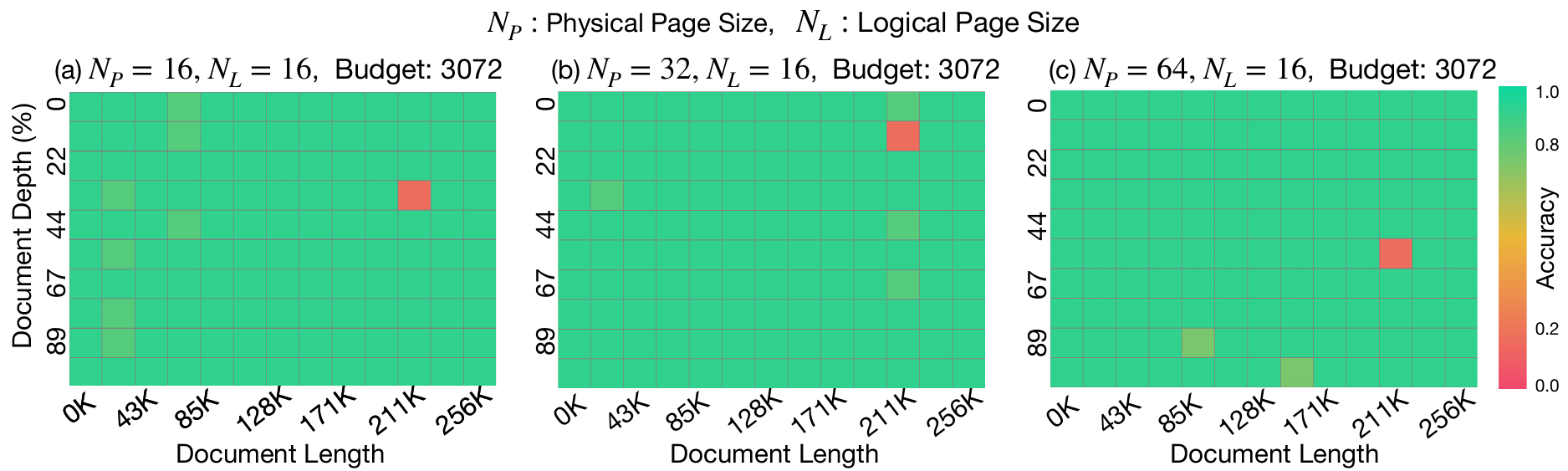}
    \caption{\textbf{Hierarchical paging} enables \system to preserve the long-context retrieval capabilities of the original model without increasing the key-value (KV) token budget. We use Llama-3-8B for the ablation.}

    \label{fig:ana:our_larger_page}
\end{figure}

\begin{figure}[t]
    \centering
    \includegraphics[width=\linewidth]{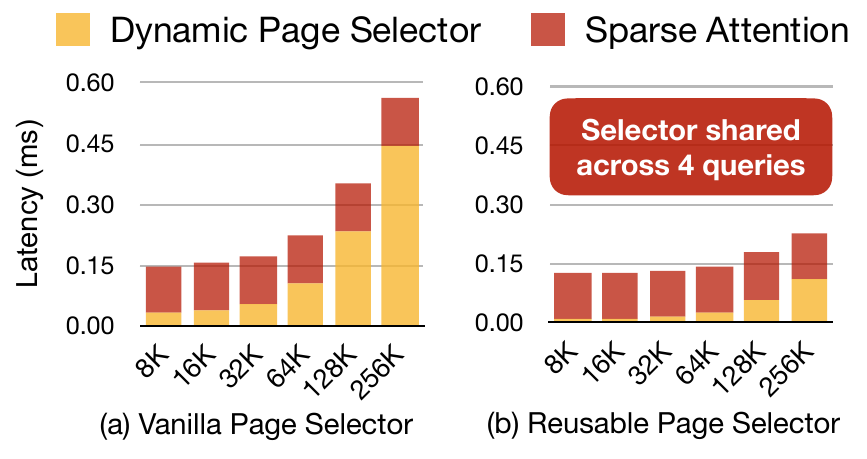}
    \caption{\textbf{Effect of Reusable Page Selection}. The overhead of the dynamic page selector is significant, as its complexity increases linearly with input sequence length. Our \textit{Reusable Page Selection} effectively mitigates this issue. The latency breakdown is evaluated on Llama-3-8B.} 
    \label{fig:ana:selector_overhead}
    \vspace{4pt}
\end{figure}

In this section, we present in-depth analysis on our design choices in the \system system from both the accuracy and the efficiency perspective. We also scrutinize the sources of performance gains in \sect{sect:results}.

\subsection{Prefilling Stage Sparse Attention Kernel}

We benchmark the performance of our block sparse attention kernel for the prefilling stage in Figure~\ref{fig:ana:prefilling_attention}. Compared with the implementation by MInference~\cite{jiang2024minference}, our kernel consistently achieves 1.3$\times$ speedup at the same sparsity level. Oracle stands for the theoretical upper-bound speedup ratio: $\text{Latency}_{\text{oracle}} = \text{Latency}_{\text{dense}} * (1-\text{sparsity})$.

\subsection{Effectiveness of Hierarchical Paging}

We use the Needle-in-a-Haystack ~\cite{LLMTest_NeedleInAHaystack} test to demonstrate that the hierarchical paging design effectively maintains the model's long-context capability on larger page blocks without increasing the token budget. In contrast to the performance drop observed with increased page granularity in Figure~\ref{fig:ana:naive-larger-page}, \system leverages a hierarchical page structure to decouple the pruning algorithm’s page granularity from the physical memory layout of the KV cache. This approach enables our sparse attention mechanism to remain both accurate and hardware-efficient. Figure~\ref{fig:ana:our_larger_page} highlights this improvement: with a page size of 64 and the same token budget, \system achieves accuracy comparable to the baseline algorithm~\cite{tang2024quest}, which prunes history tokens at a granularity of 16.

\subsection{Mitigating Page Selection Overhead}

\begin{table}[t]
\centering
\caption{The reusable page selector in \system preserves the model's long-context accuracy while significantly reducing selection overhead by \textbf{4$\times$} with a reuse interval of 4. We evaluate Llama-3-8B on RULER~\cite{nvidia_ruler} at a sequence length of 64K. LServe-$N$ denotes that the token budget for dynamic sparsity is $N$.}

\footnotesize
\scalebox{0.95}{
\begin{tabular}{ccccccc}

\toprule
Reuse Interval & Dense    & 1    & 2    & 4    & 8    & 16   \\ 
\midrule
LServe-4096 & 86.8 & 86.2 & 85.6 & 85.6 & 84.8 & 83.2 \\ 
\midrule			
LServe-8192 & 86.8 & 86.1 & 85.8 & 85.5 & 85.6 & 84.8\\ 
\bottomrule
\end{tabular}
}
\label{tab:ana:reusable_accuracy}
\vspace{10pt}
\end{table}

\paragraph{Reusable Page Selection.} During decoding, although the attention kernel maintains constant complexity due to a capped number of historical KV tokens, the complexity of the page selector still scales linearly with sequence length. As illustrated in Figure~\ref{fig:ana:selector_overhead}, for a sequence length of 128K and a 4K token budget for sparse attention, the page selector (0.24 ms) is already twice as slow as the sparse attention kernel (0.12 ms). With our reusable page selector, however, \system significantly reduces page selection overhead by a factor of $C$, where $C$ is the reuse interval. We further show that \system is resilient to different reuse interval choices. Table~\ref{tab:ana:reusable_accuracy} demonstrates no significant performance degradation until the reuse interval exceeds 8, so we set it to 4 by default in \system.

\paragraph{Context Pooling Overhead.} To enable page selection during decoding, we must calculate representative features using min-max pooling in the prefilling stage. It is important to note that a single pooling kernel executes under \textbf{1 ms}, while the entire prefilling stage completes in approximately 17 seconds with 128K context length. Consequently, the context pooling overhead is negligible.

\subsection{Sparse Attention Kernel for Decoding Stage}

\begin{figure}[t]
    \centering
    \includegraphics[width=0.9\linewidth]{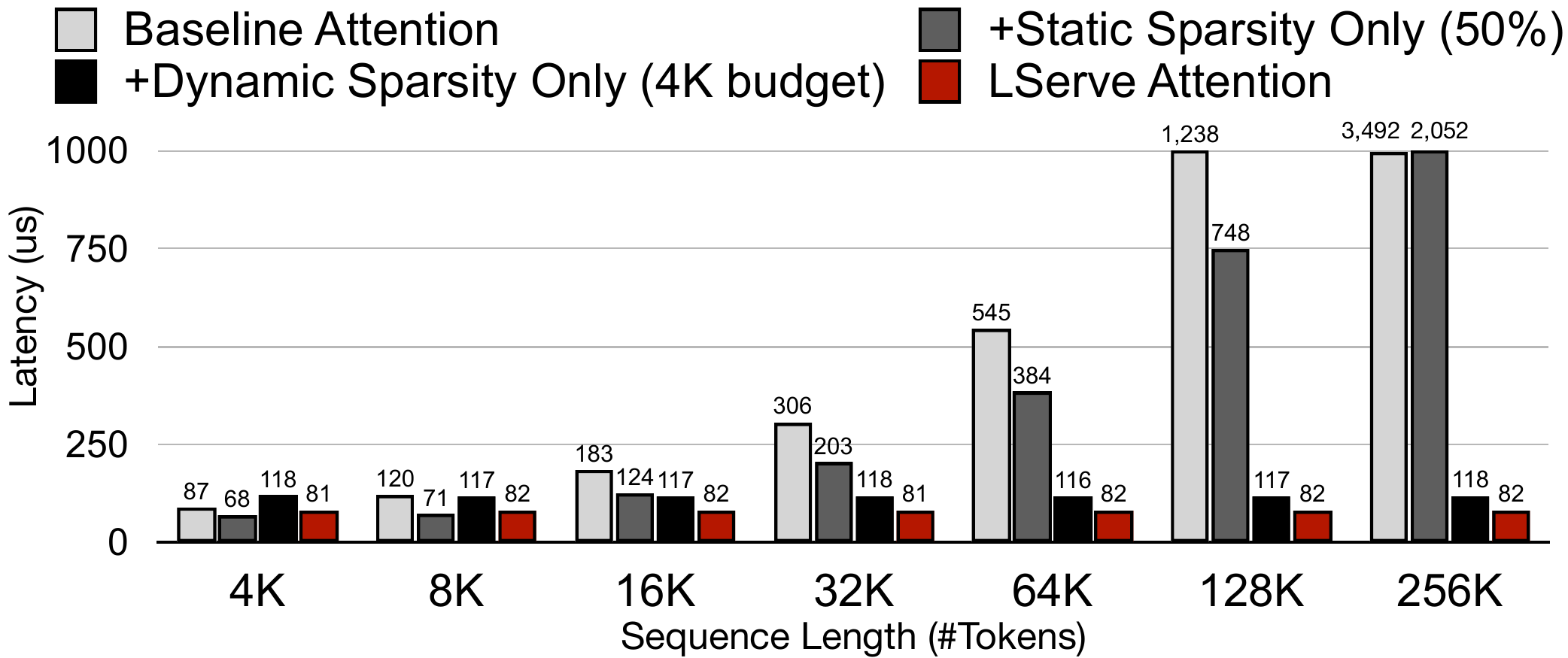}
    \caption{\textbf{Efficiency gains from static and dynamic sparsity in \system}. These sparsity patterns contribute to a compound speedup effect, with static sparsity being more effective at shorter contexts, and dynamic sparsity offering greater benefits at longer contexts. We report the latency of a single attention layer in Llama-2-7B.}
    \label{fig:ana:decoding_attn_kernel}
\end{figure}

We analyze the effectiveness of different sparsity patterns in decoding attention. In Figure~\ref{fig:ana:decoding_attn_kernel}, we apply \textit{static} sparsity by converting 50\% of attention heads to streaming heads, achieving a \textbf{1.3-1.7$\times$} speedup across various input sequence lengths. Additionally, we introduce dynamic sparsity with a fixed KV budget of 4096 tokens, which bounds the computational complexity of decoding attention to a \textbf{constant}, delivering a \textbf{30$\times$} speedup over the dense baseline for an input length of 256K.  Although sparsity selection introduces minor overhead for shorter sequences, this is mitigated by reusable page selection. Additionally, we also perform the end-to-end ablation study in Section \ref{sect:End-to-End Ablation}.

\vspace{-7pt}
\subsection{End-to-End Speedup Breakdown}
\label{sect:End-to-End Ablation}

\begin{figure}[t]
    \centering
    \includegraphics[width=0.9\linewidth]{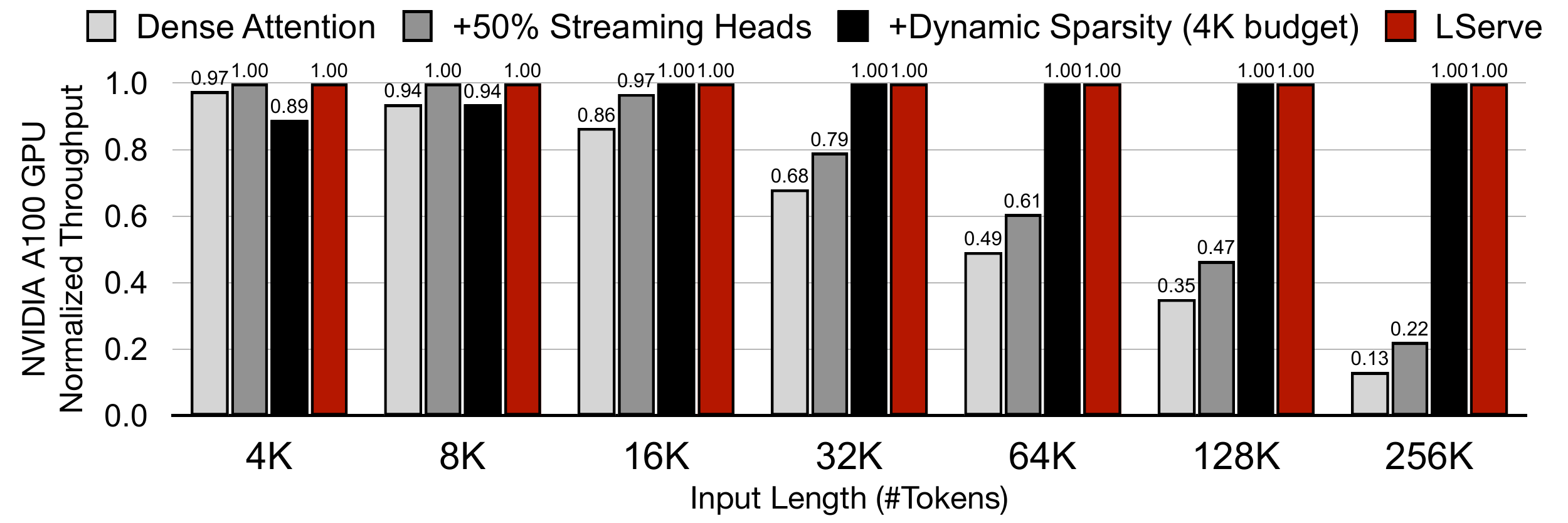}
    \caption{\textbf{End-to-end speedup breakdown in \system}: Consistent with findings from attention layer analysis, static sparsity (50\% streaming heads) yields greater benefits at shorter context lengths. In contrast, dynamic sparsity achieves up to \textbf{4.5$\times$} end-to-end speedup for longer sequences. Results are based on measurements using Llama-3-8B with unit batch size.}
    \label{fig:ana:case_study}
\end{figure}

In Figure~\ref{fig:ana:case_study}, we highlight the sources of performance improvement in \system. By leveraging static sparsity, \system achieves end-to-end speedups of up to \textbf{1.7$\times$} over the dense baseline. Additionally, dynamic sparsity, aided by a reusable page selector, significantly reduces generation latency, yielding a \textbf{7.7$\times$} speedup for sequence lengths of 256K. Lastly, \system configures sparse patterns through offline profiling, effectively avoiding slowdowns from dynamic sparsity at shorter context lengths.

\vspace{-10pt}
\section{Related Work}
\label{sect:related}

\textbf{LLM Serving Systems}. Various systems have been developed to enhance LLM deployment efficiency. Orca~\cite{orca} uses iteration-level scheduling and selective batching for distributed systems. vLLM~\cite{vllm} introduces PagedAttention, inspired by virtual memory, to optimize KV cache management. TensorRT-LLM~\cite{trtllm} is the industry’s leading solution also featuring in-flight batching and PagedAttention inspired by vLLM. LightLLM~\cite{lightllm} further reduces memory waste in PagedAttention by introducing TokenAttention. SGLang~\cite{sglang} advances LLM programming with a domain-specific language and RadixAttention. LMDeploy~\cite{lmdeploy} improves deployment with persistent batching and blocked KV cache. Nanoflow~\cite{zhu2024nanoflow} features intra-device scheduling and asynchronous CPU scheduling, while QServe~\cite{lin2024qserve} improves LLM serving throughput through \texttt{W4A8KV4} quantization and system codesign. MLC-LLM~\cite{mlcllm} accelerates deployment on edge devices via compiler-based optimizations. Inspired by contextual sparsity~\cite{liu2023deja}, PowerInfer~\cite{song2023powerinfer,xue2024powerinfer} deploys LLMs on memory-constrained devices via offloading.

\textbf{Sparse Attention}. BigBird~\cite{zaheer2020big} reduces attention complexity by blending local, global, and random attention masks. Subsequent methods like StreamingLLM~\cite{xiao2023efficient}, H2O~\cite{zhang2024h2o}, and TOVA~\cite{oren2024transformers} simplify attention patterns by discarding KV caches mid-way through the context. However, these approaches struggle to retain the original models' long-context capabilities due to limited global context modeling. Recent works like DuoAttention~\cite{xiao2024duoattention} and RetrievalAttention~\cite{liu2024retrievalattention} address this issue by introducing retrieval heads~\cite{wu2024retrieval} or combining full attention with local attention heads. SeerAttention~\cite{gao2024seerattention} introduces a learnable gate to identify block-level attention sparsity. Quest~\cite{tang2024quest} applies dynamic, query-aware sparsity for accelerated decoding, while MInference~\cite{jiang2024minference} extends similar ideas to the prefilling stage. FastGen~\cite{ge2023model} optimizes decoding by profiling attention heads to discard tokens. PQCache~\cite{zhang2024pqcache} and ShadowKV~\cite{sun2024shadowkv} further advance the selective attention methods with product quantization and low-rank decomposition. Additionally, LongLoRA~\cite{chen2023longlora} finetunes short-context LLMs to long-context ones after converting global attention to shifted sparse attention.

\vspace{-10pt}
\section{Conclusion}
\label{sect:conclusion}

We introduce \system, an efficient serving system for long-sequence LLMs that leverages hybrid sparse attention. By incorporating \textit{unified block sparse attention}, we achieve significant acceleration of the attention mechanism for both prefilling and decoding stages in long-sequence models. We further show that head-level static sparsity and query-aware dynamic sparsity are orthogonal and can be effectively combined with minimal impact on accuracy. \system surpasses state-of-the-art systems, delivering an average of \textbf{1.3$\times$-2.1$\times$} speedup in the decoding stage and up to \textbf{2.9$\times$} speedup in the prefilling stage, preserving the models' capabilities to effectively handle long-context documents and perform complex reasoning tasks.

\vspace{-10pt}

\section*{Acknowledgements}
We thank MIT-IBM Watson AI Lab, MIT AI Hardware Program, MIT Amazon Science Hub, and National Science Foundation for supporting this research. We also thank June Yang, Bo Li, and Kaiyu Xie for their helpful discussions.

\newpage

\nocite{langley00}

\bibliography{refs}
\bibliographystyle{mlsys2025}

\appendix

\clearpage

\section{Artifact Appendix}

\subsection{Abstract}

This artifact contains necessary scripts and dependencies
to faithfully reproduce the crucial experiments presented in the paper. To successfully run the experiments, a host system with x86\_64 CPUs is required, along with at least one A100 or L40S NVIDIA GPU. We also provide a pre-built docker image to simplify the environment setup process.

\subsection{Artifact check-list (meta-information)}

{\small
\begin{itemize}
  \item {\bf Program:} Accuracy evaluation and efficiency benchmarking code for LServe; efficiency benchmarking code for baseline systems such as vLLM.
  \item {\bf Compilation:} Completed in the docker.
  \item {\bf Transformations:} N/A. 
  \item {\bf Binary:} N/A.
  \item {\bf Model:} We provide a quantized version of Llama-3-8B to simplify the evaluation process. 
  \item {\bf Data set:} Included in the docker image.
  \item {\bf Run-time environment:} NVIDIA Container Toolkit \texttt{(nvidia-docker)}.
  \item {\bf Hardware:} A host with x86\_64 CPUs and at least one NVIDIA A100 GPU (recommended) or L40S GPU.
  \item {\bf Run-time state:} N/A.
  \item {\bf Execution:} All benchmarks are executed on NVIDIA GPUs, while some data pre-processing code is executed on the host CPU;
  \item {\bf Metrics:} Long-context benchmarks accuracy; LLM generation throughput.
  \item {\bf Output:} Accuracy numbers and generation throughput (tokens/second).
  \item {\bf Experiments:} Inference speed measurement for LServe and baseline systems such as vLLM; accuracy evaluation of LServe.
  \item {\bf How much disk space required (approximately)?:} 128G.
  \item {\bf How much time is needed to prepare workflow (approximately)?:} Around 1 hour to pull docker images depending on the Internet connection and CPU performance.
  \item {\bf How much time is needed to complete experiments (approximately)?:} Around 2 hours to finish the efficiency benchmarks; and 1 hour to finish the accuracy benchmarks depending on the GPU performance and number of benchmark subsets to evaluate. %
  \item {\bf Publicly available?:} Yes.
  \item {\bf Code licenses (if publicly available)?:} Apache License 2.0.
  \item {\bf Data licenses (if publicly available)?:} MIT.
  \item {\bf Workflow framework used?:} Docker.
  \item {\bf Archived (provide DOI)?:} \url{https://doi.org/10.5281/zenodo.14989916}
\end{itemize}

\subsection{Description}

\subsubsection{How delivered} %

We will provide AE reviewers with a pre-built docker image containing LServe, vLLM, and all necessary dependencies.

\subsubsection{Hardware dependencies} %

A host machine with x86\_64 CPUs and at least one NVIDIA A100 GPU.

\subsubsection{Software dependencies} 

A GPU-compatible Docker runtime environment is required.

\subsubsection{Data sets}

We require datasets such as \texttt{2wikimqa}, \texttt{dureader}, and \texttt{hotpotqa} for LongBench evaluation. All datasets are publicly available and included in the docker image.

\subsection{Installation}

We recommend that users utilize our pre-built Docker images to set up the environment and run all experiments within the GPU-supported Docker container.

\begin{lstlisting}[style=mystyle, language=bash]
docker run --gpus all -it --workdir /workspace/projects shang12138/lserve-mlsys25-ae
\end{lstlisting}

\subsection{Experiment workflow}

\subsubsection{LServe Accuracy Evaluation}

We provide push-button solution for evaluating LServe on LongBench.

\begin{lstlisting}[style=mystyle, language=bash]
cd /workspace/projects/omniserve
bash eval/scripts/LongBench/submit_longbench_dense.sh  
    # Evaluate the baseline accuracy
bash eval/scripts/LongBench/submit_longbench_sparse.sh 
    # Evaluate LServe accuracy

# The evaluation results can be found at ./eval/LongBench/pred/<model-name>/result.json 
\end{lstlisting}

\subsubsection{Throughput Benchmark}

The generation throughputs of LServe and baseline system (i.e., vLLM) can be measured with the following commands.

\begin{lstlisting}[style=mystyle, language=bash]
# LServe benchmark
cd /workspace/projects/omniserve    
bash scripts/lserve_benchmark/launch.sh 
# Results in ./results.csv

# vLLM benchmark
cd /workspace/projects/vllm   
bash launch_server.sh # Start vLLM server
# When the vLLM server has been launched,
# start a new terminal in the same docker
cd /workspace/projects/vllm 
bash run_vllm.sh # Launch evaluation
# Results in ./results.csv
\end{lstlisting}

\subsection{Evaluation and expected result}

We provide reference numbers for evaluation results in this section. Please note that absolute speed measurements may vary slightly, even on identical GPU platforms, due to differences in machine conditions. However, the relative acceleration ratios should remain consistent. Additionally, due to randomness inherent in LLM generation, accuracy results might show minor deviations from the reported reference numbers.

\vspace{15pt}

\begin{table}[h]
    \centering
    \small
        \centering
        \begin{tabular}{cccc}
            \hline
            Sequence Length & vLLM (ms) & LServe (ms) & Speedup \\
            \hline
            64k  & 12.51  & 11.49 & 1.09x \\
            96k  & 14.49  & 12.05 & 1.20x \\
            128k & 16.34  & 12.74 & 1.28x \\
            160k & 18.20  & 12.88 & 1.41x \\
            192k & 21.73  & 13.30 & 1.63x \\
            224k & 21.96  & 13.73 & 1.60x \\
            256k & 23.72  & 14.20 & 1.67x \\
            320k & 27.45  & 15.10 & 1.82x \\
            \hline
        \end{tabular}
        \caption{Generation latency of LServe and baseline (vLLM).}
        \label{tab:decoding_comparison}
\end{table}

\vspace{15pt}

\begin{table}[h]
    \centering
    \begin{tabular}{ccccc}
        \hline
        Model & \multicolumn{2}{c}{Llama-3-8B} \\
        \hline
        Benchmark & Dense & \system \\
        \hline
        2WikiMQA  & 26.2 & 27.0 \\ 
        DuReader  & 22.3 & 25.6 \\ 
        HotpotQA  & 41.1 & 40.8 \\ 
        MultiNews & 27.6 & 27.1 \\ 
        Qasper    & 29.1 & 28.5 \\ 
        \hline
    \end{tabular}
    \caption{LongBench evaluation results.}
    \label{tab:benchmark_comparison}
\end{table}

\vspace{15pt}

\subsection{Methodology}

Submission, reviewing and badging methodology:

\begin{itemize}
  \item \url{http://cTuning.org/ae/submission-20190109.html}
  \item \url{http://cTuning.org/ae/reviewing-20190109.html}
  \item \url{https://www.acm.org/publications/policies/artifact-review-badging}
\end{itemize}

\end{document}